\newif\ifarxiv % -- choose preprint (arxiv) version versus conference format
\arxivtrue

\newif\ifreview % -- turn on and off author info, page+line numbers and appendix proofs
\reviewfalse

\newif\iffinal % -- turn on and off colored comments
\finaltrue

\documentclass{article}

\ifarxiv\else
\ifreview
\usepackage[nonatbib,main]{neurips_2026}
\else
\usepackage[nonatbib,main,final]{neurips_2026}
\fi
\fi

\usepackage[utf8]{inputenc}
\usepackage[utf8]{inputenc}

\ifarxiv
\usepackage[a4paper,left=2cm,right=2cm,top=2cm,bottom=2cm]{geometry}
\usepackage[onehalfspacing]{setspace}
\usepackage{amsthm}
\usepackage{authblk}
\fi

\usepackage[hidelinks,bookmarks]{hyperref}
\usepackage[numbers,sort]{natbib}
\usepackage{amsmath,amssymb,amsfonts,mathtools,dsfont}
\usepackage{nicefrac} 

\usepackage{booktabs} % spacing and styling
\usepackage{xspace}
\usepackage{graphicx}
\usepackage{xspace}
\usepackage{url}
\usepackage{enumitem}

\usepackage{caption} % captions ans subfigures
\usepackage{subcaption}

\usepackage{booktabs}
\usepackage{svg}

\usepackage{float}

\usepackage[algo2e,ruled,vlined,linesnumbered]{algorithm2e} % algorithm type setting
\SetAlgoSkip{}
\DontPrintSemicolon

\ifarxiv
\usepackage[appendix=inline]{apxproof} % no appendix proof
\else
\usepackage[appendix=append]{apxproof}  % having appendix proof
\fi

\allowdisplaybreaks[3] % break equations between pages

\usepackage{tikz}
\usetikzlibrary{shapes,arrows}
\usetikzlibrary{decorations}
\usetikzlibrary{decorations.markings}
\usetikzlibrary{plotmarks}
\usetikzlibrary{calc}
\usetikzlibrary{mindmap}
\usetikzlibrary{shadows}
\usetikzlibrary{backgrounds}
\usetikzlibrary{patterns}
\usetikzlibrary{shapes.symbols}

\usepackage{pgfplots}
\usepackage{pgfplotstable}
\usepgfplotslibrary{statistics}
\pgfplotsset{compat=1.18}

\ifarxiv
\newtheorem{theorem}             {Theorem}
\newtheorem{lemma}      [theorem]{Lemma}

\fi

\newtheoremrep{theorem}{Theorem}
\newtheoremrep{lemma}     [theorem]{Lemma}
\newtheoremrep{definition}[theorem]{Definition}
\newtheoremrep{conjecture}[theorem]{Conjecture}
\newtheoremrep{corollary} [theorem]{Corollary}
\newtheoremrep{proposition}[theorem]{Proposition}

\usepackage{tabularx,colortbl}

\makeatletter
\g@addto@macro\bfseries{\boldmath}
\makeatother

\newcommand{\ie}{i.\,e.\xspace}
\newcommand{\wrt}{w.\,r.\,t.\xspace}
\newcommand{\eg}{e.\,g.\xspace}

\newcommand{\wlo}{w.\,l.\,o.\,g.\xspace}

\newcommand{\N}{\mathbb{N}}

\DeclareMathOperator*{\logplus}{log^+}
\DeclareMathOperator*{\logmult}{log^{\times}}

\DeclareMathOperator{\Unif}{Unif}                         % uniform distribution
\newcommand*{\bigO}{\mathcal{O}}

\newcommand{\prob}[1]{\Pr\left(#1\right)}                 % probability
\newcommand{\searchspace}{\ensuremath{\mathcal{X}}\xspace}       % search space of valid programs

\newcommand{\maxproblemfull}{$\textsc{Max}$-depth-$D$-$\{\times,+\}$-$\{t\}$\xspace} % the MAX problem
\newcommand{\maxproblemparam}[3]{$\textsc{Max}$-depth-#1-#2-#3\xspace}               % parametrised form
\newcommand{\maxproblem}{$\textsc{Max}$\xspace}                                      % short name

\newcommand{\Nin}{\ensuremath{N_{\mathrm{in}}}}                       % number of input nodes
\newcommand{\Nout}{\ensuremath{N_{\mathrm{out}}}}                      % number of output nodes
\newcommand{\Nfunc}{\ensuremath{N_{\mathrm{func}}}}                     % number of function nodes
\newcommand{\Nall}{\ensuremath{N}}                        % total number of nodes
\newcommand{\Kfunc}{\ensuremath{K}}                       % cardinality of the function lookup table
\newcommand{\arity}{\ensuremath{k}}                       % arity of the functions
\newcommand{\genomelength}{\ensuremath{L}}                   % the final genome length

\newcommand{\mut}{\ensuremath{\mathrm{mut}}}                % a mutation operator
\newcommand{\pmut}{\ensuremath{\mathrm{pmut}}}              % transition matrix for mutation
\newcommand{\pmutsam}{\ensuremath{\pmut_{\mathrm{SAM}}}}    % for SAM
\newcommand{\pmutprob}{\ensuremath{\pmut_{\mathrm{PROB}}}}  % for PROB
\newcommand{\pmutsamlb}{\ensuremath{\pmut_{\mathrm{SAM}}}'} % lower bound estimate for SAM 

\newcommand{\fblkina}[1]{\ensuremath{#1.\mathrm{in}_1}\xspace}  % node that input-1 of a function node points to
\newcommand{\fblkinb}[1]{\ensuremath{#1.\mathrm{in}_2}\xspace}  % node that input-2 points to
\newcommand{\fblkfunc}[1]{\ensuremath{#1.f\xspace}}             % function chosen from the lookup table
\newcommand{\fblkval}[1]{\ensuremath{\mathrm{val}(#1)\xspace}}  % output value of a node
\newcommand{\fblkdom}[1]{\ensuremath{\mathcal{V}(#1)\xspace}}   % domain of the output value

\newcommand{\fblkvalparam}[2]{\ensuremath{\mathrm{val}_{#2}(#1)\xspace}}  % output value of a node, version where the function evaluation is specified

\newcommand{\toggleplot}[1]{{\textcolor{red}{Plots removed to increase compilation speed. Use command $\backslash$toggleplot in preamble to reinsert them.}}}

\iffinal
\newcommand{\andre}[1]{}
\newcommand{\romann}[1]{}
\newcommand{\cuong}[1]{}
\newcommand{\newedit}[1]{#1}
\newcommand*{\todo}[1]{}
\else
\newcommand{\andre}[1]{\textcolor{green!50!black}{[(Andre) #1]}}
\newcommand{\romann}[1]{\textcolor{purple}{[(Roman) #1]}}
\newcommand{\cuong}[1]{\textcolor{orange}{[(Cuong) #1]}}
\newcommand{\newedit}[1]{\textcolor{blue}{#1}}
\newcommand*{\todo}[1]{\textcolor{red}{\textrm{(TODO: #1)}}}
\fi

\ifarxiv
\else
\fi

\ifarxiv
\else
\fi

\begin{document}

\ifarxiv
\author[1]{Duc-Cuong~Dang}
\author[2]{Roman~Kalkreuth}
\author[1]{Andre~Opris}

\affil[1]{University of Passau, Passau, Germany}
\affil[2]{RWTH Aachen University, Aachen, Germany}
\date{}
\else
\author{
    Duc-Cuong~Dang\\
    Chair of Algorithms for Intelligent Systems,\\
    University of Passau, Passau, Germany\\
    \texttt{duccuong.dang@uni-passau.de}
\And
    Roman~Kalkreuth\\
    Chair for AI Methodology (AIM),\\ 
    Faculty of Computer Science,\\
    RWTH~Aachen~University, Aachen, Germany\\
    \texttt{kalkreuth@aim.rwth-aachen.de}
\And
    Andre~Opris\\
    Chair of Algorithms for Intelligent Systems,\\
    University of Passau, Passau, Germany\\
    \texttt{andre.opris@uni-passau.de}
}
\fi

%\title{Cartesian Genetic Programming is Efficient on the \textsc{Max} Problem}
%\title{A Proven Exponential Speedup for Cartesian Genetic Programming on the \textsc{Max} Problem}
%\title{Cartesian Genetic Programming on the \textsc{Max} Problem: A Proven Exponential Speedup}
%\title{Analysis of Cartesian Genetic Programming on the \textsc{Max} Problem: A Proven Exponential Speedup}
\title{Runtime Analysis of Cartesian Genetic Programming on \textsc{Max}: \ifarxiv\\\fi 
       A Proven Exponential Speedup}
\maketitle

\begin{abstract}
Genetic Programming (GP) is a search paradigm inspired from natural evolution for the automated discovery of expressions, functions, and computer
programs. Cartesian GP (CGP) is a flavor of GP that uses a graph-based model to
encode candidate programs, as contrasted to the conventional Tree-based GP
(TGP). Since its inception two decades ago, CGP has predominantly been analyzed
empirically, thus little is known regarding its theoretical performance guarantees.

This paper analyzes CGP for the \textsc{Max} problem, which has been only
rigorously studied for TGP with a proven expected runtime superlinear in the
optimal program length $n$. We prove that the expected time for 
the ($1$+$1$)~CGP employing common mutation operators to evolve an equivalent
optimal of the same output is polylogarithmic in $n$, specifically
at most $\bigO(\log^9{n})$. This is remarkable, as it shows an
exponential speedup by switching to CGP.
%% -- Is this the ChatGPT version of the one above? It removed the style of the original writing, added some confusion to the first sentence
%This paper analyzes CGP for the Max problem, which has so far only been rigorously studied for TGP, where the expected runtime is known to be superlinear in the optimal program length $n$. We show that the expected time for the ($1$+$1$)~CGP algorithm, using standard mutation operators, to evolve an equivalent optimal program with the same output is polylogarithmic in $n$, specifically at most $\bigO(\log^9 n)$. This represents an exponential speedup compared to TGP, achieved by switching to CGP.
%
Our results shed light on the benefit and the compactness of the graph-based
representation for programs, and on how CGP can navigate its
search and fitness spaces efficiently. This is a first proven performance
guarantee for CGP on an established benchmark problem. Experiments
complement our theoretical findings.
%\ifarxiv\else
%\keywords{Genetic Programming, Runtime Analysis, Graph Representation}
%\fi
\end{abstract}

\section{Introduction}\label{sec:intro}

Genetic Programming (GP) is a search paradigm and a subfield of
Evolutionary Computation (EC) and Artificial Intelligence (AI) that evolves
computer programs. The goal of applying GP is to automate the process
of discovering computer programs based solely on given specifications, such as
constraints and input-output examples.
Early work by Cramer~\cite{Cramer1985} and Dickmanns~et~al.~\cite{DickmannsSW1987} 
applied Genetic Algorithms (GA) to program generation. Koza~\cite{Koza1989} 
later introduced tree-based GP (TGP) to evolve LISP S-expressions, and with 
follow-up work~\cite{Koza1993} established the research field. 
In tree-based GP (TGP), or ``Koza-style GP'', programs are represented 
as trees: leaves are input variables or constants from a terminal set $T$, 
internal nodes are compute functions from a set $F$, and the root produces 
the output. Since then, GP has been extended to include different flavors, 
depending on how programs are encoded, 
\ie having different search spaces, for example 
    linear-sequence~\cite{Perkis1994,Openshaw94buildingnew}, 
    graph-based~\cite{Poli1996,Miller1999}, 
    rule-based~\cite{RyanCO1998} models, 
or on how they are evolved,
\ie using different search operators, for example 
    semantically driven designs~\cite{DBLP:conf/ppsn/MoraglioKJ12}. 
%GP has since diversified into variants with different representations, such as linear~\cite{Perkis1994,Openshaw94buildingnew}, graph based~\cite{Poli1996,Miller1999}, and rule based models~\cite{RyanCO1998}, as well as different search operators, including semantic approaches~\cite{DBLP:conf/ppsn/MoraglioKJ12}. 
%More recently, general purpose AI has %% autoencoders and transformers are not general-purpose AI 
\newedit{Recently, AI-related techniques have} 
been used to enhance GP. Components 
from Neural Architecture Search (NAS), such as transformers~\cite{DBLP:conf/gecco/AnthesSR25} 
and autoencoders~\cite{DBLP:conf/eurogp/WittenbergR23}, have been integrated 
into GP, while Large Language Models (LLMs) have been used as search operators 
for evolving program code~\cite{DBLP:journals/nature/RomeraParedesBNBKDREWFKF24, DBLP:journals/gpem/HembergMO24,novikov2025alphaevolvecodingagentscientific,lange2025shinkaevolveopenendedsampleefficientprogram}. 

Despite these recent high-profile advances, GP remains largely experimental, 
with most rigorous theoretical results confined to TGP~\cite{LissovoiO2020}. 
In particular, the tradeoffs associated with non tree-based search spaces 
are not well understood. 
\newedit{
In a survey of 
%open issues in 
GP research, O’Neill~et~al.~\cite{ONeillVGB2010} 
also identified understanding GP representations for programs, and subsequently 
finding the appropriate ones for problems, as a highly challenging 
open issue.} 
Motivated by 
%this gap, 
\newedit{these gaps,} 
we study Cartesian Genetic 
Programming (CGP), a popular and non tree-based flavor of GP introduced by 
Miller~\cite{Miller1999}. 
% 
%% -- Explain CGP and applications
CGP uses a graph-based representation with a fixed length vector encoding of 
programs (see Section~\ref{sec:def-cgp}). 
%This design offers advantages over TGP, including natural support for multiple outputs and explicit control over connectivity, which can improve search efficiency.
%In spite of being 
Although introduced two decades ago and successfully applied in many areas such as 
    NAS~\cite{10.1162/evco_a_00253}, 
    logic synthesis~\cite{DBLP:conf/ddecs/FiserSVS10,Sekanina2011}, 
    policy learning~\cite{DBLP:conf/gecco/WilsonCLM18}, 
    cryptography~\cite{10.1145/3319619.3326869}, 
and image processing with human competitive results~\cite{naturecommcgp}, very little 
is known regarding its theoretical guarantees. 
%% To our knowledge, only two works provide rigorous analysis~\cite{KalkreuthD2019,kalkreuth:thesis}.
%
%CGP adopts a graph-based representation together with a fixed-length vector encoding of programs (see Section~\ref{sec:def-cgp}). This design offers several conceptual advantages over TGP. In particular, CGP naturally supports multiple output structures, broadening its range of applications, and allows explicit control over connectivity, which can positively influence search efficiency. In spite of its inception two decades ago and of many successful 
%applications across a wide range of domains, including 
    %NAS~\cite{10.1162/evco_a_00253}, 
    %logic synthesis~\cite{DBLP:conf/ddecs/FiserSVS10, Sekanina2011}, 
    %policy learning~\cite{DBLP:conf/gecco/WilsonCLM18}, 
    %cryptography~\cite{10.1145/3319619.3326869}, 
    %and image processing %, which has also produced 
        %with human-competitive results~\cite{naturecommcgp}, 
%very little is known about its theoretical 
%performance guarantees. As far as we know, only two studies~\cite{KalkreuthD2019,kalkreuth:thesis} rigorously analyzed CGP. 

%% -- Introduce the research questions
%
%We address the following 
Our research questions are: what advantages does the graph-based 
representation of CGP provide, and how efficiently can CGP explore 
its search space for a given fitness function? 
%\todo{Cite the paper on Open Issues in Genetic Programming from Banzhaf~et~al., as here we address one of their open questions.} 
To answer these, 
we apply runtime analysis~\cite{BOOKDOERR,DoerrN2021survey,Jansen2013}. 
%, as done for TGP~\cite{LissovoiO2020}. 
Here, GP is treated as a randomized algorithm, and performance is 
measured by the expected number of fitness evaluations 
%needed 
to evolve a desired program. Efficiency corresponds to polynomial growth, 
whereas exponential growth indicates inefficiency.

%Hence, we will address the following open research questions. 
%What benefit does the graph-based representation of CGP bring? 
%How efficient is CGP in navigating its designed search space for a given fitness function? 
%To answer those, we will apply the methodology of runtime analysis~\cite{BOOKDOERR,DoerrN2021survey,Jansen2013} 
%to CGP, as it has been successfully done for TGP~\cite{LissovoiO2020}. A GP search 
%is then treated as a randomized algorithm and we analyze its random time, measured in 
%number of fitness (program) evaluations, to evolve a desired program. 
%The search is said efficient if the expected time grows polynomially to the dimension 
%of the problem. It is called inefficient if the growth is exponential. 

%\subsection{Our contributions}
\textbf{Our contribution}: 
%We conduct a first runtime analysis of CGP on the \maxproblem problem,
%specifically the \maxproblemfull, which is an established benchmark in TGP. The task is to evolve a program that produces the maximum output 
%for a bounded tree depth $D$ using the function set of only 
%addition and multiplication and the terminal set of only one real constant $t\geq 1$. 
%We first show that the optimal TGP program for this problem of size $n=2^{D+1}-1$ can be 
%efficiently encoded in CGP using a vector of integers of length linear to $D=\Theta(\log{n})$. 
%We then prove that the ($1$+$1$)~CGP equipped 
%with commonly used mutation operators, including the widely used single active-gene mutation (SAM)~\cite{GoldmanP2013},
%requires at most $\bigO(\log^9{n})$ fitness evaluations in expectation to produce
%the optimal program if $t\in[1,2)$, and this bound is reduced to $\bigO(\log^8{n})$
%if $t\geq 2$. Since it has been shown in~\cite{KoetzingSNO2014} 
%that ($1$+$1$)~TGP requires an expected time superlinear to $n$ to evolve 
%the same program, %thus 
%switching to CGP provides an exponential speedup. 
%This answers our first research question on the benefit of the graph-based 
%representation of programs that CGP brings.
We present a first runtime analysis of CGP on the \maxproblem problem, specifically \maxproblemfull, a standard TGP benchmark. The goal is to evolve a program that maximizes the output for bounded depth $D$ using only addition and multiplication, with a single constant $t \geq 1$.
We show that the optimal TGP program of size $n = 2^{D+1}-1$ can be encoded in CGP with a vector of length linear in $D = \Theta(\log n)$. We then prove that the $(1+1)$ CGP with common mutation operators, including single active gene mutation (SAM)~\cite{GoldmanP2013}, finds its optimal program in expected time\footnote{
%Time 
Here time bounds are expressed in $n$ 
%to ease the comparison with the literature results, 
for literature comparison, 
%while 
later in our results they 
will be written in $D$.}
 $\bigO(\log^9 n)$ for $t \in [1,2)$, improving to $\bigO(\log^8 n)$ for $t \geq 2$.
Since the $(1+1)$ TGP requires superlinear time in $n$ for the same task~\cite{KoetzingSNO2014}, this implies an exponential speedup. This answers our first research question on the benefit of the graph-based 
representation of programs that CGP brings. 
%Here we express time bounds as functions of $n$ to ease the comparison 
%with the existing results of TGP, while the bounds in our actual result are written 
%as functions of $D$. %so as to ease the analysis.
% 
%Our result is the first to rigorously prove an exponential 
%performance gap between CGP and TCP. It sheds the light on the benefit of 
%the graph-based representation of CGP.

Our analysis follows the progress in fitness of the evolving CGP program.
Particularly, we identify which location and elements in the vector-encoded CGP 
program that can be changed to make a significant improvement to fitness and
with a sufficient probability. This leads to the polynomial time bounds and
answers our second research question, that is, CGP can navigate its search space
efficiently by steadily increasing the fitness of the current program. Theoretical
research of EC algorithms is predominantly done on the search space of bit strings,
including recent cutting edge results,
\eg~\cite{LenglerCross,LiZZD2026,OPRIS2026}. 
%, but some like
%Qian~et~al.~\cite{QianZTY2018} have looked into integer-valued vectors with a
%specific functionality. 
Our work is the first to analyze integer-valued vectors
that encode acyclic graphs representing computer programs. In particular, the
components of the vectors are bounded, they can be active or inactive, and model
a mix of ordinal variables (encoding nodes in a chain) as well as categorical ones
(encoding indices of a lookup table, \ie see Section~\ref{sec:def-cgp}).
% 
%\ifarxiv
%In %complement 
%addition to the upper bounds, we also provide a basic lower bound of $\Omega(\log^4{n})$ 
%for the expected runtime 
%using a worst-case analysis for the initial program. 
%of a worst-case initialization.
%\fi 
Finally, %experiments are conducted for both TGP and CGP to 
\newedit{our experiments conducted on \maxproblem} 
for both TGP and CGP 
complement
the theoretical findings. 
%We proof lower and upper runtime bounds for \maxproblem and present corresponding empirical results that have been obtained in the framework of our experiments. While tree-based GP has been shown to have exponential runtime and can therefore not be considered to be efficient on \maxproblem, 

%Due to the space limit, some of our proofs are deferred to the appendix document. 

\section{Related work}\label{sec:related-work}

%Early theoretical work on GP focused on schema theory~\cite{PoliM2003,PoliM2003a,PoliMR2004}. The first runtime results for GP appeared in Neumann~et~al.~\cite{NeumannOW2011}
%and followed by~\cite{WagnerNU2015,DoerrKLL2017,KoetzingLLM2020}, analyzing the \textsc{Majority} and \textsc{Order} problems. 

%% -- State-of-the-art of runtime analysis (should this be in Related Work?)
Runtime analysis has 
%significantly 
largely 
contributed to our understanding of the performance of EC algorithms, 
%various EAs, 
\eg see~\cite{DoerrN2021survey,BOOKDOERR} for surveys 
%and books 
and~\cite{UpBian,SMSSPEED,Opris2025multimodal,LiZZD2026,OPRIS2026} for some recent progresses. It has also led to the development of new algorithms and operators that not only have proven runtime guarantees for artificial benchmarks, but also perform competitively on practical problems~\cite{LenglerCross,DoerrDE2015,DoerrKV2024,GoldmanP2015,DangELQ2022,DangEQ2024}. 
%,Doerr2019opl,Doerr2022}. %% -- they only do OM and OMM in experiments

Early theoretical work on TGP focused on schema theory~\cite{PoliM2003,PoliM2003a,PoliMR2004}. The first runtime analysis appeared in~\cite{NeumannOW2011} and followed by~\cite{WagnerNU2015,DoerrKLL2017,KoetzingLLM2020}, analyzing the \textsc{Majority} and \textsc{Order} problems to evolve structured trees (non-executable programs).  
%problems in~\cite{WagnerNU2015,DoerrKLL2017,KoetzingLLM2020}. 
% 
%This work and the follow-up~\cite{WagnerNU2015,DoerrKLL2017,KoetzingLLM2020} explored how simple TGP algorithms evolve trees 
%with 
%of specific structures. 
%The program evaluations in those problems are artificial 
%the fitness is obtained by inspecting the tree in some order instead of executing it.
%as the evolved programs are non-executable. 
%
%% -- MAX and AND (Effiency of simple TGP)
The \maxproblem problem was introduced in~\cite{GathercoleR1996} and studied in~\cite{LangdonP1997,LangdonP2002} 
%to illustrate performance issues with 
for crossover-only GP algorithms. 
Runtime results of~\maxproblem appeared in K\"{o}tzing~et~al.~\cite{KoetzingSNO2012,KoetzingSNO2014},
indeed the first to prove performance guarantees for evolving executable programs. 
They showed that the expected runtime of simple ($1$+$1$)~TGP systems using mutation operators 
based on the so-called HVL-prime operator (see Section~\ref{sec:def-max} or~\cite{NeumannOW2011}) 
is $\bigO(n\log{n})$ and $\bigO(n^2)$, and proved some benefit for biased mutations. 
%Here $n$ is the length of the optimal program,
%\ie 
%the number of nodes of a complete binary tree.
Mambrini~and~Oliveto~\cite{MambriniO2016} were the first to rigorously analyze GP 
for evolving Boolean functions. 
%problems %that resemble practical GP 
%where the fitness evaluation is based on both executing the program and 
%varying the inputs. 
%By assuming that all inputs can be evaluated and the function 
%set $F$ with only the necessary binary gate, 
They showed that the same TGP systems 
%as~\cite{NeumannOW2011,KoetzingSNO2014} are efficient in evolving the $\mathrm{AND}_n$ 
%function but are not for $\mathrm{XOR}_n$ (the logical $\mathrm{AND}$ and $\mathrm{XOR}$
%respectively of $n$ inputs). 
are efficient for $\mathrm{AND}_n$ but not for $\mathrm{XOR}_n$, reflecting
the fact that $\mathrm{XOR}_n$  
%The later is the consequence of a well-known complexity 
%result that 
%the parity function 
%$\mathrm{XOR}_n$ 
is \newedit{PAC-learnable~\cite{Valiant1984} but not evolvable~\cite{Valiant2009}}. 
%They also analyzed the settings of not evaluating all inputs, and of 
%$\mathrm{AND}_n$ with negated inputs in the terminal set $T$. 
%Their work spawned several follow-up papers (\eg~\cite{DoerrLO2019,LissovoiO2018}) 
%where more realistic assumptions were considered.
%
%% ChatGPT gebbrish
%The $\maxproblem$ problem was introduced in~\cite{GathercoleR1996} and empirically studied in~\cite{LangdonP1997,LangdonP2002} 
%%to highlight limitations of 
%for crossover-only TGP. The first runtime results were obtained by K"otzing et al.~\cite{KoetzingSNO2012,KoetzingSNO2014}, who gave the first performance guarantees for evolving executable programs. They showed that simple $(1+1)$ TGP using HVL prime mutation~\cite{NeumannOW2011} runs in expected time $\bigO(n \log n)$ and $\bigO(n^2)$, and identified cases where biased mutation improves performance. Here $n$ is the size of the optimal program, i.e., the number of nodes in a complete binary tree.
%Mambrini and Oliveto~\cite{MambriniO2016} were the first to rigorously analyze GP on problems with explicit input sampling. Under assumptions allowing evaluation of %all inputs and a restricted function set, they showed that the same TGP variants are efficient for $\mathrm{AND}_n$ but inefficient for $\mathrm{XOR}_n$, reflecting %the fact that $\mathrm{XOR}_n$ is not PAC learnable~\cite{Valiant1984}. They also studied variants with partial input evaluation and with negated inputs for %$\mathrm{AND}_n$. 
% 
These results motivated further work under more realistic settings~\cite{DoerrLO2019,LissovoiO2018}.

%% -- CGP
%While runtime analysis has been successfully conducted for TGP, not many studies have been done for the 
%other flavors of GP. 
%other search spaces of programs.  
Mathematical analysis of GP on the other search spaces of programs is scarce.
The only studies that we know of are Kalkreuth and Droschinsky~\cite{KalkreuthD2019, kalkreuth:thesis}, 
which proposed the first rigorous analysis for the $(1+1)$-CGP 
with SAM mutation. They showed an expected runtime bound 
$\Theta(n \log{n})$ for the \newedit{\textsc{Sum}} problem, and bounds $\Omega(n^2)$ and 
$\bigO(n^2 \log{n})$ for the \newedit{\textsc{CountingOperators}} problem.
%two problems \textsc{Sum} and \textsc{CountingOperators} on $n$ constant inputs. 
%Expected runtime bounds of $\Theta(n \log{n})$ were proven for \textsc{Sum} and $\Omega(n^2)$ and $\bigO(n^2 \log{n})$ for \textsc{CountingOperators}  by respectively using drift analysis~\cite{DoerrJW2010}. 
%, an arithmetic test problem and COUNTING OPERATORS, a combinatorial counting problem, which have been proposed to analyze the efficiency of CGP's abilities to evolve connection properties that can often be found in the arithmetic or logical domain. 
%However the work is limited in a sense that merely one function node was considered for the analysis and therefore focused primarily on specific properties of the mutation operator in order to establish required connections to input nodes which does not allow more general conclusions to be drawn about the search performance of GGP. 
However, only one computing node was defined in those problems; thus it is unclear how the results 
%are scaled with a large number of nodes. 
can be generalized for a large number of nodes. %and other mutations.
% 
%% -- Integer-value problems
% Besides rather poor theoretical insight, empirical studies on CGP induced search led to a considerable 
CGP programs are encoded as a vector of integers of 
%increasing ranges.
diverse ranges. The rigorous analysis of Evolutionary Algorithms (EAs) on integer-valued 
problems was initiated by 
    Doerr~et~al.~\cite{DoerrJS2011}, and was further explored 
    in~\cite{DoerrP2012} on the benefit of high mutation rates, %the benefit of non-uniform distributions in mutation, 
    in~\cite{DoerrDK2015,DoerrDK2018} for self-adaptation, and 
    in~\cite{QianZTY2018} for the multiset selection problem.
Analysis of unbounded integer-valued problems was also considered, \eg~\cite{Doerr_Krejca_Rudolph_2025,Rudolph1994}.
%However, it 
It remains open how the insights gained from those studies can be useful 
for CGP where the vector encodes a graph.

% -- Algorithmic problems on graphs
Theoretical work on EC algorithms for 
%algorithmic 
problems on graphs can be found %for example 
in~\cite{BossekNPS2021,BhuvaDHS2022,DangN0OS2025,DoerrHK2012,DoerrJKNT2013,SudholtT2012}.
%
% -- Neural networks, and neural architecture search
The original form of CGP was inspired by FPGA circuits~\cite{Miller1999} and proposed to use a 2D grid topology 
for the graphs with a level-back parameter, 
%to control the connectivity, 
thus sharing some similarities with Neural Networks (NNs). However, in NNs the graph topology and functionality on nodes are fixed, then back propagation and numerical optimization are used to compute the edge weights~\cite{Cybenko1989,Hornik1991}. In contrast, CGP %on the other hand 
optimizes both the topology and functionality on nodes in parallel using evolutionary search. Recent works on NNs and graph learning can be found in~\cite{FroeseHN2022,HertrichS2023,XuHLJ2019,LuoLL0W2025}.
There has been a recent interest in runtime analysis of Evolutionary NAS (ENAS), 
% for classification problems, 
\ie see Fischer~et~al.~\cite{FischerLW2023,FischerWW2024} and Lv~et~al.~\cite{Lv0002024,LvQYS2024,Lv0LF02025}. 

%\romann{I will address recent approaches here, predominately evolutionary/LLM-driven coding agents}
%\cuong{OK, but note that we should focus more on theory papers for those. We can ask AI for the most important ones and read the introduction those and that should give the others.}
%
%Because of a lack of theoretical development, empirical studies have been the main gateway for the CGP community to gain 
%some levels of understanding of CGP characteristics, \ie 
%    redundancy and efficiency~\cite{DBLP:journals/tec/MillerS06}, 
    %bloat~\cite{DBLP:conf/eurogp/TurnerM14}, and 
    %the role of 
    %genetic drift~\cite{DBLP:journals/gpem/TurnerM15a}.
% 
%% Additional representations
%Besides theoretical work on TGP and CGP, the first rigorous theoretical analysis of the generalization performance of Geometric Semantic Genetic Programming (GSGP)~\cite{DBLP:conf/ppsn/MoraglioKJ12} system has recently been proposed by Corus et al.~\cite{CorusO2024}.
% 
% 
%Another flavour of GP that has been theoretically explored is Geometric Semantic GP (GSGP)~(\eg see \cite{DBLP:conf/ppsn/MoraglioKJ12,CorusO2024}), 
%and in the broader field of program synthesis, runtime analysis has been used for showing the benefit of lexicase selection~\cite{MetevierSS2019} in~\cite{HelmuthLL2022}. 

\section{Cartesian Genetic Programming}\label{sec:def-cgp}

\begin{figure}[hb]\centering
\begin{tikzpicture}[scale=\ifarxiv1\else0.9\fi,
every node/.style={scale=0.9}]\small
\newcommand\nodeindex[1]{\scriptsize\textcolor{blue}{#1}}
\tikzset{middlearrow/.style={
                decoration={markings,
                mark= at position 0.5 with {\arrow{#1}} ,
            }, postaction={decorate} } }
\tikzstyle{block}=[rectangle, draw, rounded corners, minimum height=0.6cm]
\tikzstyle{node}=[circle, draw, fill=white, minimum size=0.7cm]
\tikzstyle{edge}=[]

\begin{scope}[shift={(-3.5,2.5)}] %[shift={(0,4.6)}] % lookup table
\node[anchor=west] (lt) at (-1.0,1.2) {\textbf{lookup table}};
\node[below of=lt, node distance=3.5em] {\footnotesize
\begin{tabular}{c|c|l}
index & symbol & func.\\ \hline
$0$ & $+$ & addition \\
$1$ & $-$ & subtraction \\
$2$ & $\times$ & multiplication \\
\end{tabular}
};
\end{scope}

\begin{scope}[shift={(0,2.5)}] % phenotype
\node[anchor=west] at (-0.8,1.2) {\textbf{phenotype}};
\node[node,rectangle,label=-180:{\nodeindex{$0$}}] (n0) at (0,0.5)   {$i_1$};
\node[node,rectangle,label=-180:{\nodeindex{$1$}}] (n1) at (0,-0.5)   {$i_2$};
\node[node,label=90:{\nodeindex{$2$}}]  (n2) at (1.5,0) {$+$};
\node[node,label=90:{\nodeindex{$3$}}]  (n3) at (3.0,0) {$\times$};
\node[node,label=90:{\nodeindex{$4$}}] (n4) at (4.5,0) {$-$};
\node[node,label=-90:{\nodeindex{$5$}}] (n5) at (6.0,0) {$\times$};
\node[node,diamond,label=90:{\nodeindex{$6$}}] (n6) at (7.5,0) {$o_1$};

\draw[edge] (n0) edge[bend left,middlearrow={>}]  (n2);
\draw[edge] (n0) edge[bend right,middlearrow={>}] (n2);
\draw[edge] (n2) edge[bend left,middlearrow={>}]  (n3);
\draw[edge] (n1) edge[bend right,middlearrow={>}] (n3);
\draw[edge] (n3) edge[bend left,middlearrow={>}]  (n4);
\draw[edge] (n2) edge[bend right,middlearrow={>}] (n4);
\draw[edge] (n0) edge[bend left,middlearrow={>}]  (n5);
\draw[edge] (n3) edge[bend right,middlearrow={>}] (n5);
\draw[edge] (n4) edge[bend left,middlearrow={>}]  (n6);
\end{scope}

\begin{scope}[shift={(0,0.6)}] % genotype
\node[anchor=west] (nnl) at (-0.8,0) {\textbf{genotype}};
\node[block,right of=nnl,label=90:{\nodeindex{$2$}},node distance=5.2em] (nn2) {$0$ \nodeindex{$0$} \nodeindex{$0$}};
\node[block,right of=nn2,label=90:{\nodeindex{$3$}}] (nn3) {$2$ \nodeindex{$2$} \nodeindex{$1$}};
\node[block,right of=nn3,label=90:{\nodeindex{$4$}}] (nn4) {$1$ \nodeindex{$3$} \nodeindex{$2$}};
\node[block,right of=nn4,label=90:{\nodeindex{$5$}}] (nn5) {$2$ \nodeindex{$0$} \nodeindex{$3$}};
\node[block,right of=nn5,label=90:{\nodeindex{$6$}},node distance=0.8cm] (nn6) {\nodeindex{$4$}};
\end{scope}
\end{tikzpicture}
\caption{Example of a CGP program that computes the expression $o_1=(i_1+i_1)\times i_2 - (i_1+i_1)$.
         Node~5 is a so-called inactive node because no paths connect it to the output.}
\label{fig:example-cgp-program}
\end{figure}
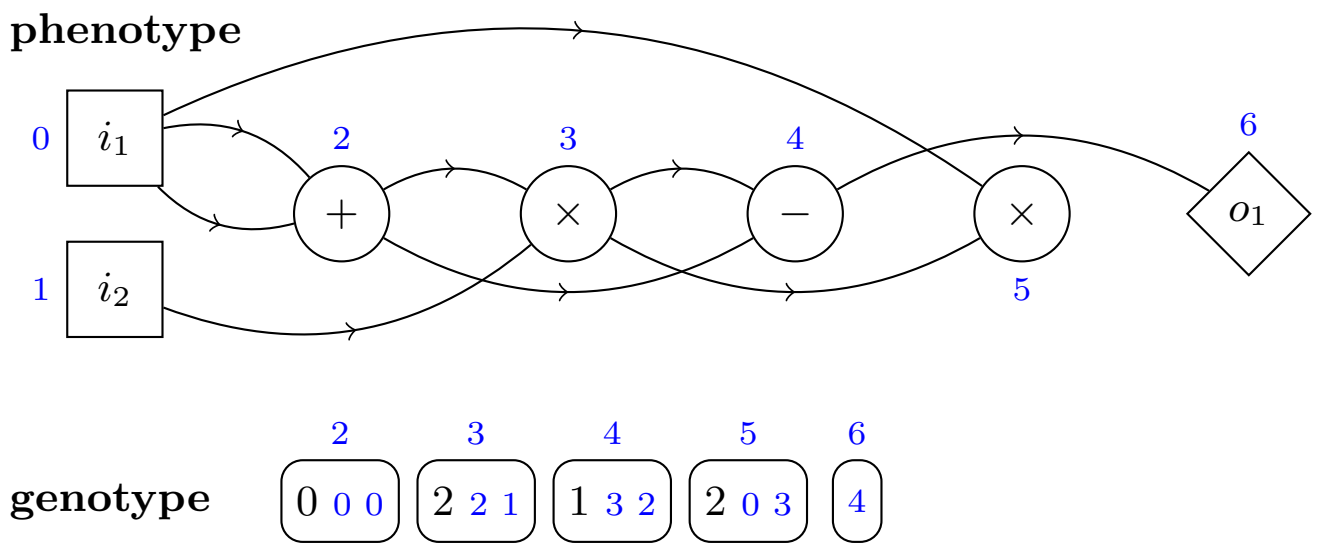

CGP is a popular representative of graph-based GP. It is defined for 
    $\Nin$ inputs, $\Nout$ outputs, and $\Nfunc$ (internal) function nodes 
and uses a lookup table of 
    $\Kfunc$ functions 
    %that are indexed from $0$ to $\Kfunc-1$
    %the arity of each function is at 
    of arity most $\arity$. 
A (valid) program for such a CGP is represented by a directed acyclic graph $G=(V,E)$,
called the \emph{phenotype}, that tells how the outputs are computed 
from the inputs 
%via function composition (see Figure~\ref{fig:example-cgp-program}). 
by composing lookup functions (see Figure~\ref{fig:example-cgp-program}).
Inputs and the lookup table correspond to the terminal set $T$ and function set $F$ in TGP.
%by composing 
%the functions defined in the lookup table and the inputs, 
%\eg see Figure~\ref{fig:example-cgp-program} for an example.
%The inputs and the lookup table model the same concept as the terminal set $T$ and
%the function set $F$ in TGP.

%The original CGP~\cite{Miller1999} 
%put the function nodes on a 
%2D grid,  
%however 
We follow the modern approach for CGP in which 
%they 
function nodes are arranged on one line. 
Nodes are indexed by $V = [\Nall-1]_0$, where $\Nall := \Nin + \Nfunc + \Nout$, ordered as inputs, function nodes, then outputs. 
Here $[n]$ for $n\in N$ is the notation for $\{1,\dots,n\}$ and $[n]:=[n]\cup\{0\}$. 
Input nodes have no incoming edges, while each function node has $\arity$ incoming edges 
from nodes of smaller index, encoding ordered arguments (some edges are ignored if the 
arity of the lookup function is smaller than $k$). Each output node has one incoming edge from a non output node.
%Thus we set $V=[\Nall-1]_0$ 
%where $\Nall:=\Nin + \Nfunc + \Nout$ 
%\ie nodes are indexed from $0$ to $N-1$, starting with 
%the input nodes, %$\Nin$, 
%then the function nodes %, %$\Nfunc$, 
%and 
%the output ones.
%Input nodes have no incoming edges (connections) while each output node has exactly 
%one coming from a non-output node.  
%Each function node is labeled with a function in the lookup table and has exactly
%$\arity$ incoming edges from nodes of strictly smaller indices, thus from the inputs
%or from the outputs of the preceding function nodes. These edges represent the sequence of 
%arguments that the 
%function in the 
%lookup table takes, thus they follow an order.
%If the arity of 
%that function is smaller than $k$, then only some 
%first edges in the order are taken into account. 

The graph $G$ is encoded as a vector of $\Nfunc(\arity+1)+\Nout$ integers, 
called the \emph{genotype}. The vector is divided into blocks indexed from 
$\Nin$ to $\Nall-1$, matching 
the indices of the nodes in phenotype that they encode. A block with index $i<\Nin+\Nfunc$ (a function block) contains
$\arity+1$ integers, the first, 
called a \emph{function gene}, is an integer %a number 
in $[\Kfunc-1]_0$ 
(an index in the lookup table) while the rest, 
called \emph{connection genes}, are integers in $[i-1]_{0}$. 
A block of larger index $i$ (an output block) only contains 
an integer in 
$[\Nall-\Nout-1]_0$, 
called an \emph{output connection gene}. 
The set $\searchspace$ of all genotypes (all valid CGP programs)
is a space of integer-valued vectors.

We consider the simplest CGP algorithm given in 
Algorithm~\ref{algo:one-plus-one-cgp}, the so-called ($1$+$1$)~CGP. Starting 
from a first program initialized uniformly at random, in each iteration a new 
$y$ offspring program is generated by mutating the current one. The offspring 
will replace the parent if it has better or equally good fitness. Here $f(x)$ 
denotes the fitness of a program $x$, 
which is problem-dependent and assumed to be given. 
By convention in runtime analysis, we let the algorithm run infinitely 
then estimate the expected time to reach some target state. 
Practical CGP algorithms 
%often 
only use a slightly more general framework, 
%which a variant
%of the ($1$+$\lambda$)~EA and $\lambda=1$
%, it results in 
%is exactly our algorithm. %,
%and we leave the generalisation for any $\lambda$ to future work.
%where 
where $\lambda$ (typically $4$) offspring programs like $y$ are 
generated per iteration.

\begin{minipage}[t]{0.34\textwidth}\null
\begin{algorithm2e}[H]\small
%\begin{algorithm2e}[ht]\small 
$x\sim \Unif(\searchspace)$\;
\For{$\mathrm{iter}:= 1 \to \infty$}{
    $y:=\mathrm{mutate}(x)$\;
    \If{$f(y)\geq f(x)$}{
        $x:=y$%\;
    } 
}
\caption{The ($1$+$1$)-CGP on the space $\searchspace$ of programs.}
\label{algo:one-plus-one-cgp}
\end{algorithm2e}
\end{minipage}\hfill
\begin{minipage}[t]{0.62\textwidth}\null
\begin{algorithm2e}[H]\small
%\begin{algorithm2e}[ht]\small
$b:=\lfloor i/(\arity+1)\rfloor$ \tcp{\scriptsize block index}
$j:=i \mod (\arity+1)$ \tcp{\scriptsize index inside a block}
%$u:=x[i]$ \tcp{\footnotesize old value}
\eIf{$i<\Nfunc(\arity+1)$}{
    \lIf{$j=0$}{
        $S:=[\Kfunc-1]_0$ \tcp*[h]{\scriptsize valid function genes}
    }
    \lElse{
        $S:=[\Nin+b-1]_0$ \tcp*[h]{\scriptsize valid input connection genes}
    }
}{
    $S:=[\Nin+\Nfunc-1]_0$ \tcp{\scriptsize valid output connection genes}
}
\If{$|S\setminus\{x[i]\}|\geq 1$}{
    $S:=S\setminus\{x[i]\}$ \tcp{\scriptsize avoid the old value if possible}
}
%$v\sim \Unif(S)$\;
\KwRet{$v\sim \Unif(S)$}
\caption{$\textsc{NewValue}(x, i)$.}
\label{algo:new-value}
\end{algorithm2e}
\end{minipage}

%\subsection{CGP Mutations and a Transfer Lemma}
\textbf{CGP mutations:}\label{sec:cgp-mutations} %
Mutation operators can 
%decide 
\newedit{choose} 
to change any gene (component) of the genotype 
(vector) to a new value uniformly at random while respecting the constraint on their range. 
Such a change 
%this 
is done 
%with the help of 
by the procedure \textsc{NewValue} which is 
%summarized 
shown in Algorithm~\ref{algo:new-value}.
%When enforcing the change is not possible, \ie the range only contains one integer, 
%%for example if the program
%%has only one input then the first function node can only take this input as 
%%its arguments, 
%the old value is returned instead.

%Two mutation operators are often considered in CGP, the first one is the so-called 
%\emph{probabilistic mutation}, that is, each gene has the same probability $p$ to be 
%modified. This operator, as shown in Algorithm~\ref{algo:mut-pointwise}, is 
%the generalization of the standard bit mutation on bit strings 
%to the space $\searchspace$ of CGP programs. 
%The other operator, shown in Algorithm~\ref{algo:mut-sam}, is the \emph{single 
%active-gene mutation} (SAM)~\cite{GoldmanP2013} which repeatedly modifies genes chosen 
%uniformly at random until one associated with a so-called \emph{active} node is modified. 
%A node is active if there is a path from it to one of the outputs in the phenotype 
%representation (thus output nodes are active by this definition). Since the detection 
%of active nodes can be done during the program evaluation, we can then assume that 
%there is a function $\textsc{IsActive}$ that can tell if a given gene belongs to 
%an active node or not in an evaluated program. 

Two mutation operators are %commonly used in 
common in CGP. The first, probabilistic mutation (Algorithm~\ref{algo:mut-pointwise}), mutates each gene independently with probability $p$, generalizing standard bit mutation on bit strings to the CGP search space $\searchspace$. The second, single active gene mutation (SAM)~\cite{GoldmanP2013} (Algorithm~\ref{algo:mut-sam}), 
repeatedly \newedit{mutates genes chosen} uniformly at random until one associated with a so-called \emph{active} node is modified. A node is active if it lies on a path to an output \newedit{node} in the phenotype\newedit{. All output nodes are therefore} active by \newedit{this} definition. 
%Since active nodes can be identified during evaluation, we assume access to a function $\textsc{IsActive}$ that determines whether a gene belongs to an active node.
Active nodes can be identified during evaluation, thus we assume \newedit{having} access to an $\textsc{IsActive}$ function. 

\begin{minipage}[t]{0.46\textwidth}\null
\begin{algorithm2e}[H]\small
%\begin{algorithm2e}[ht]\small
Copy $x$ to $y$\;
\For{$i:= 0 \to \Nfunc(\arity+1)+\Nout-1$}{ 
    $r \sim \Unif((0,1))$\;
    \lIf{$r<p$}{
        $y[i]:=\textsc{NewValue}(y, i)$%\;
    }
}
\KwRet{$y$}
\caption{$\textsc{MutationPROB}(x, p)$.}
\label{algo:mut-pointwise}
\end{algorithm2e}
\end{minipage}
\hfill
\begin{minipage}[t]{0.5\textwidth}\null
\begin{algorithm2e}[H]\small %[ht]\small
%\begin{algorithm2e}[ht]\small
Copy $x$ to $y$\;
$\mathrm{active} := \mathrm{false}$\;
\While{$\neg \mathrm{active}$}{
    $i\sim\Unif([\Nfunc(\arity+1)+\Nout-1]_0)$\;
    $y[i]:=\textsc{NewValue}(y, i)$\;
    $\mathrm{active} := \textsc{IsActive}(x,i) \wedge (y[i]\neq x[i])$\;
}
\KwRet{$y$}
\caption{$\textsc{MutationSAM}(x)$.}
\label{algo:mut-sam}
\end{algorithm2e}
\end{minipage}

%In our analysis, we 
\textbf{A transfer lemma:} % 
We will focus on \newedit{SAM mutation}
as it is parameter-free and is
used more often in practice.
However, upper bound results are asymptotically transferable to the 
%probabilistic mutation 
other \newedit{operator} %mutation 
as follows.
%this is a \emph{global operator}. 
%
Formally, a mutation operator $\mut$ is described by a transition matrix $\pmut$ where 
each entry is $\pmut(x,y)\coloneqq\prob{\mut(x)=y}$ for a pair $x,y\in \searchspace$ 
of programs. Thus we have the matrices $\pmutsam$ and $\pmutprob$ respectively. 
The probabilistic mutation is a so-called \emph{global mutation} since
%It holds that
$\pmutprob(x,y)>0$ for all $x,y\in\searchspace$.
%, \ie it is always possible to reach any program from another. 
In contrast for SAM mutation, $\pmutsam(x,y)>0$ requires that $y$ modifies 
exactly one active node of $x$  
and possibly some inactive ones.  
%, \ie this is a global mutation operator. 

For $x, y$ satisfying $\pmutsam(x,y)>0$, it is often non-trivial to compute 
$\pmutsam(x,y)$. But we can estimate a lower bound $\pmutsamlb(x,y)$ for it, which 
is defined as follows. Let $\genomelength \coloneqq \Nfunc(\arity+1)+\Nout$ be the genome length, 
$M$ be the set of genes where $x$ and $y$ differ, and 
for each $s\in M$, let $q(s)$ be the probability that $\textsc{NewValue}(x,s)$ returns $y[s]$. 
We then fix a specific order of the genes of $M$ for the while-loop in SAM to 
%follow 
modify in $x$ so that the gene associated with the active node is modified last.
This order corresponds to one specific way to reach $y$ from $x$, and since
in each iteration of the loop, a gene is selected uniformly at random (probability $1/L$) 
%to get a new value 
to be modified, 
%by calling $\textsc{NewValue}$, 
thus %it holds that 
$\pmutsam(x,y)%$ is bounded from below by 
%$
\geq \prod_{s\in M} (1/L)q(s) \eqqcolon  \pmutsamlb(x,y)$.
%This is a lower bound 
%The inequality is  
%because we have not taken into account the other $(|M|-1)!-1$ 
%equivalent orders of $M$, and also other ways to reach $y$. 
%because we only consider one 
%(short) 
%specific way to reach $y$. 
For $x,y$ where $\pmutsam(x,y)=0$, we 
define $\pmutsamlb(x,y) \coloneqq 0$. 
The following lemma, allowing the transfer of results, shows that 
%that under an identical start, if the ($1$+$1$)~CGP using 
%SAM mutation can reach a target program $x_k$ by following a specific path, then the same 
%algorithm with the probabilistic mutation can also reach $x_k$ using that path. 
the ($1$+$1$)~CGP using the probabilistic mutation can follow the same path
as the algorithm using SAM mutation under the same initialization. 
%
%In particular, 
The probabilities of these two 
%events 
walks are asymptotically equal if that of the former event 
was estimated using $\pmutsamlb$ and the path length is constant. 
%Thus, the lemma allows the transfer of results.

\begin{lemmarep}\label{lem:from-SAM-to-PROB}
Let $L:=\Nfunc(\arity+1)+\Nout$. %be the genome length of CGP programs. 
Consider a sequence of $k+1$ CGP programs $x_0,\ldots,x_{k}$ such that for all $i\in[k]$, 
%we have 
    $f(x_{i-1})\leq f(x_{i})$ %, 
    %$H(x_{i-1},x_{i})=O(1)$ %% no need as we omit the factorial factor in the lower bound
    and $\pmutsamlb(x_{i-1},x_{i})\geq p_i >0$. 
Then after $k$ iterations 
the ($1$+$1$)~CGP using SAM mutation and initialized with program $x_0$ has probability
of at least $\tilde{p}:=\prod_{i=1}^{k}p_i$ to change its current program to $x_k$. Moreover,
the ($1$+$1$)~CGP using the probabilistic mutation with parameter $p=\frac{1}{L}$
under the same condition can reach program $x_k$ with probability 
at least $\tilde{p}\cdot e^{-k}$,
which is $\Omega(\tilde{p})$ if $k=O(1)$.
%\andre{is this something like $\tilde{p}/e^k$? Maybe we should add that, and then say if $k=O(1)$...} 
%\cuong{We already said this in ``Moreover,$\ldots$''.}
%\andre{maybe one thing one has to be careful about is, that SAM can deal with "repetitions": it can touch certain outputs more than one time and probabilisstic mutation not. Maybe one has to exclude this event also from the statement.} 
%\cuong{Yes, that is why there is no mention of $\pmutsam(x,y)$ in the lemma. There is no repetition in $\pmutsamlb(x,y)$.} 
Here the asymptotic notation is 
%with respect to 
\wrt 
$L$. 
\end{lemmarep}
\begin{appendixproof}
Since $p_i>0$ for all $i\in[k]$, all the transitions from $x_{i-1}$ to $x_i$ are 
feasible for SAM mutation, that is, only one active node is changed per transition. 
The statement for \newedit{SAM} 
then holds because the sequence of search points has 
non-decreasing fitness, thus the target solutions will be accepted once they are created,
and moreover the operator is applied independently in $k$ iterations. 

%\cuong{Fixed the typo issues, and added index $i$ for $M$ below for clarification.}

For the second statement,
let $M_i$ the set of genes where $x_{i-1}$ and $x_i$ differ, we note that $|M_i|\geq 1$. Suppose 
now that the algorithm starts with solution $x_{i-1}$ and uses the probabilistic mutation instead, 
then each gene of $M_i$ can be changed with the same probability $p=1/L$ while the others remain 
unchanged with probability $(1-1/L)^{L-|M_i|}$. 
Thus we get 
%\andre{In the calculation below, there is a symbol $q(s)$. How is its relationship to the statement of the lemma? I saw it is defined as "returns value of $q[s]$"!}
%\cuong{All symbols are explained in the paper.} 

\begin{align*}
&\pmutprob(x_{i-1},x_{i})\\ 
  &= \left(1-\frac{1}{L}\right)^{L-|M_i|}\prod_{s\in M_i} \frac{q(s)}{L}
   = \left(1-\frac{1}{L}\right)^{L-|M_i|} \pmutsamlb(x_{i-1},x_{i})\\
  &\geq \left(1-\frac{1}{L}\right)^{L-1} \pmutsamlb(x_{i-1},x_{i})
   \geq \frac{\pmutsamlb(x_{i-1},x_{i})}{e}.
\end{align*}
Here we use \newedit{the definition of $\pmutsamlb(x_{i-1},x_{i})$ as given before the lemma 
and inequality} $(1-1/L)^{L-1}\geq 1/e$. Since this holds for all $i\in[k]$, we deduce that 
the probabilistic mutation can follow the same path as SAM starting from $x_0$ to reach $x_k$ 
with a probability of at least $\tilde{p}/e^k$. The proof is completed by noting that $k=O(1)$.
\end{appendixproof}

\begin{toappendix}
The factor $1/e^k$ in the proof is pessimistic because the probabilistic mutation does not 
need to follow the same path as SAM, \eg it can be achieved by 
a direct jump based on the \newedit{Hamming distance} $H(x_0,x_k)$ 
\newedit{between the encoded programs $x_0$ and $x_k$.}
%\andre{This is not defined in the preliminaries!}
%\cuong{Good point, it is now explained on the spot.} 
However, it is sufficient for our purpose of \newedit{transferring asymptotic results}. 
\end{toappendix}

%\subsection{Compactness of CGP Programs}
\textbf{Compactness of CGP programs:}\label{sec:compact-prog} %
A prerequisite for the efficiency of CGP is that its programs should not be too large. 
%This is based on the fact that, unlike in TGP programs, a CGP program can re-use its sub-structures.
Unlike in TGP, where a value computed at a node can only be fed to one parent, in CGP, 
the value can be fed to any subsequent nodes. 
%Thus sub-structures can be re-used. 
Therefore, a CGP program can inherently leverage reusable subprograms.
This fact is well-known in the GP community~\cite{MillerT2000}, and we introduce the following 
lemma to formalize the idea.

\begin{lemmarep}\label{lem:cgp-compactness}
For every TGP program $x$
that uses functions from 
%the function 
a set $F$ for its internal nodes $I(x)$ and elements of
%the terminal
a terminal set $T$ for its leaves, there exists an equivalent 
CGP program $z$ that produces the same output as $x$ and uses $T$ inputs, $F$ in the 
lookup table, and at most $|I(x)|$ function nodes. Moreover, if there exist $y_1,y_2\in I(x)$ 
such that the subtrees rooted at these nodes are disjoint and identical (reusable subprograms), 
then $z$ only requires at most $|I(x)| - |I(y_1)|$ function nodes.
\end{lemmarep}
\begin{appendixproof}
To build $z$ in CGP, it is clear that we only need \newedit{to put the functions of}
$F$ in the lookup table, and \newedit{to set the elements of} $T$ 
%for the inputs as the elements of $T$ can be used 
%in any of the function nodes. 
as the inputs. 
%, and the same holds for the function set $F$.  
%\andre{I could not follow the last sentence.} 
%\cuong{Clarified.} 
We then label every node $i \in I(x)$ by its depth $d(i)$ in $x$, and put all the nodes of $I(x)$ 
as the function nodes of $z$ in a non-increasing order of $d(i)$, \ie nodes with larger $d(i)$ 
have smaller indices in $z$ (and ties are broken arbitrarily). The program $z$ is completed by 
connecting the output to the last function node (with the largest index), and clearly no more 
than $|I(x)|$ function nodes are used. 

Consider that there are disjoint and identical subtrees rooted at $y_1$ and $y_2$ in $x$. 
Assume \wlo that $d(y_2)\geq d(y_1)$, then we can delete all nodes of $I(y_1)$ in 
$z$ and redirect all the severed inputs of nodes $s$ that were connected to a node $r\in I(y_1)$
to the corresponding node $r'\in I(y_2)$. This is possible without re-ordering the nodes in $z$ 
because $d(s)\leq d(y_1)-1 < d(y_2)$ thus all nodes $s$ that need to take the output value 
of $y_1$ as an argument are located after $y_2$ in $z$, and since the subtrees $y_1$ and $y_2$ 
are identical the node $r'$ always exists. Moreover, it follows from the disjointness of the 
subtrees that the output value of $y_2$ remains unchanged and equal to that of $y_1$ after the 
removal of $I(y_1)$. This reduces the required number of function nodes by $|I(y_1)|=|I(y_2)|$ 
without changing the output of $z$.
\end{appendixproof}

\section{The Max Problem}\label{sec:def-max}

The $\maxproblem$ problem was introduced in~\cite{GathercoleR1996}, empirically studied in~\cite{LangdonP1997,LangdonP2002}, and analyzed theoretically in~\cite{KoetzingSNO2012,KoetzingSNO2014}. In \maxproblemparam{$D$}{$F$}{$T$}, the goal is to construct a program of maximum output value using a function set $F$ of binary functions, a terminal set $T$ of constants, and a maximum tree depth of $D$. A \newedit{valid} TGP program has at most $2^D-1$ internal nodes and $2^D$ leaves, hence at most $n := 2^{D+1}-1$ nodes in total.
The analysis of \maxproblem in~\cite{KoetzingSNO2012,KoetzingSNO2014} 
%(also covering \maxproblemlesser, which we do not consider here) 
studied mutation-only TGPs based on the HVL-prime operator~\cite{NeumannOW2011}. Each application \newedit{of HVL-prime on a TGP program} chooses uniformly at random one of three operations \newedit{to execute}: (i) substitute a function node, (ii) insert a function at a leaf, or (iii) delete a subtree.
In the ($1$+$1$) TGP-single variant, HVL-prime is applied once per mutation; in ($1$+$1$) TGP-multi, it is applied $k$ times, where $k \sim \mathrm{Poisson}(1)$. \newedit{The ($1$+$1$) search strategy is Algorithm~\ref{algo:one-plus-one-cgp} but on the search space $\mathcal{X}$ of TGP programs, and all trees with depth exceeding $D$ have fitness $-\infty$.} The authors of~\cite{KoetzingSNO2014} derived the following 
%upper 
bounds on the optimization time, and a matching lower bound of $\Omega(n \log n)$ follows via a coupon collector argument.

\begin{theorem}[Theorems~5, 6, 7 and~10 in~\cite{KoetzingSNO2014}]\label{thm:tgp-max}
The expected optimization time of the ($1$+$1$)~TGP-single algorithm on 
\maxproblemfull is $\bigO(n\log{n})$ for all $t>0$, while that of ($1$+$1$)~TGP-multi on 
\maxproblemparam{$D$}{$\{\times,+\}$}{$\{1\}$} is $\bigO(n^2)$. %Here $n$ is $2^{D+1}-1$.
\end{theorem}

\begin{figure}[ht]\centering
\begin{tikzpicture}[scale=\ifarxiv1\else0.9\fi,
every node/.style={scale=0.9}]\small
\newcommand\nodeindex[1]{\scriptsize\textcolor{blue}{#1}}
\tikzset{middlearrow/.style={
                decoration={markings,
                mark= at position 0.5 with {\arrow{#1}} ,
            }, postaction={decorate} } }
\tikzstyle{block}=[rectangle, draw, rounded corners, minimum height=0.6cm]
\tikzstyle{node}=[circle, draw, fill=white, minimum size=0.7cm]
\tikzstyle{edge}=[]
\tikzstyle{tgpleaf}=[rectangle,draw,fill=white]
\tikzstyle{tgpnode}=[circle,draw,fill=white]
\tikzstyle{tgproot}=[diamond,draw,fill=white]

\begin{scope}[shift={(0.0,3.4)}] % lookup table
\node[anchor=west] (lt) at (-0.5,0.0) {\textbf{lookup table}};
\node[right of=lt,node distance=12em] {\footnotesize
\begin{tabular}{c|c|l}
index & symbol & func.\\ \hline
$0$ & $+$ & addition \\
$1$ & $\times$ & multiplication \\
\end{tabular}
};
\end{scope}

\begin{scope}[shift={(0.0,1.8)}] % phenotype
\node[anchor=west] at (-0.5,0.8) {\textbf{phenotype}};
\node[node,rectangle,label=-180:{\nodeindex{$0$}}] (n0) at (0,0)    {$t$};
%\node[node,label=-180:{\nodeindex{$0$}}] (n0) at (0,-0.5) {$t$};
%\node[node,label=-180:{\nodeindex{$1$}}] (n1) at (0,0.5)  {$0$};
\node[node,label=90:{\nodeindex{$1$}}]   (n2) at (1.5,0)  {$+$};
\node[node,label=90:{\nodeindex{$2$}}]   (n3) at (3.0,0)  {$\times$};
\node[node,label=90:{\nodeindex{$3$}}]   (n4) at (4.5,0)  {$\times$};
\node[node,label=90:{\nodeindex{$4$}}]   (n5) at (6.0,0)  {$\times$};
\node[node,diamond,label=90:{\nodeindex{$5$}}]   (n6) at (7.5,0)  {$o_1$};

\draw[edge] (n0) edge[bend left,middlearrow={>}]  (n2);
\draw[edge] (n0) edge[bend right,middlearrow={>}] (n2);
\draw[edge] (n2) edge[bend left,middlearrow={>}]  (n3);
\draw[edge] (n2) edge[bend right,middlearrow={>}] (n3);
\draw[edge] (n3) edge[bend left,middlearrow={>}]  (n4);
\draw[edge] (n3) edge[bend right,middlearrow={>}] (n4);
\draw[edge] (n4) edge[bend left,middlearrow={>}]  (n5);
\draw[edge] (n4) edge[bend right,middlearrow={>}] (n5);
\draw[edge] (n5) edge[bend left,middlearrow={>}]  (n6);
\end{scope}

\begin{scope}[shift={(0.0,0.4)}] % genotype
\node[anchor=west] (nnl) at (-0.5,0) {\textbf{genotype}};
\node[block,right of=nnl,label=90:{\nodeindex{$1$}},node distance=6.5em] (nn2) {$0$ \nodeindex{$0$} \nodeindex{$0$}};
\node[block,right of=nn2,label=90:{\nodeindex{$2$}}] (nn3) {$1$ \nodeindex{$1$} \nodeindex{$1$}};
\node[block,right of=nn3,label=90:{\nodeindex{$3$}}] (nn4) {$1$ \nodeindex{$2$} \nodeindex{$2$}};
\node[block,right of=nn4,label=90:{\nodeindex{$4$}}] (nn5) {$1$ \nodeindex{$3$} \nodeindex{$3$}};
\node[block,right of=nn5,label=90:{\nodeindex{$5$}},node distance=0.8cm] (nn6) {\nodeindex{$4$}};
\end{scope}

\begin{scope}[shift={(8.5,3.4)},scale=0.8,every node/.style={scale=0.8}]\scriptsize % equivalent TGP program
\node[anchor=west] at (1.3,0.8) {\small \textbf{the equivalent TGP program}};
\foreach \i [evaluate=\i as \x using \i / 2] in {0,...,15} { % leaves
    \node[tgpleaf] (l\i) at (\x,-4) {$t$};
}
\foreach \i [evaluate=\i as \x using 0.2+\i] in {0,...,0} { % nodes at depth 3
    \node[tgpnode] (d3\i) at (\x,-3) {$+$};
}
\foreach \i [evaluate=\i as \x using 0.2+\i] in {1,...,1} { % nodes at depth 3
    \node[tgpnode,fill=black!10!white] (d3\i) at (\x,-3) {$+$};
}
\foreach \i [evaluate=\i as \x using 0.2+\i] in {2,...,3} { % nodes at depth 3
    \node[tgpnode,fill=black!30!white] (d3\i) at (\x,-3) {$+$};
}
\foreach \i [evaluate=\i as \x using 0.2+\i] in {4,...,7} { % nodes at depth 3
    \node[tgpnode,fill=black!50!white] (d3\i) at (\x,-3) {$+$};
}
\foreach \i [evaluate=\i as \cl using 2*\i, % drawing edges in between
             evaluate=\i as \cr using 2*\i + 1] in {0,...,7} {
    \draw (d3\i) -- (l\cl.-270);
    \draw (d3\i) -- (l\cr.-270);
}
\foreach \i [evaluate=\i as \x using 0.65 + 2*\i] in {0,...,0} { % nodes at depth 2
    \node[tgpnode] (d2\i) at (\x,-2) {$\times$};
}
\foreach \i [evaluate=\i as \x using 0.65 + 2*\i] in {1,...,1} { % nodes at depth 2
    \node[tgpnode,fill=black!30!white] (d2\i) at (\x,-2) {$\times$};
}
\foreach \i [evaluate=\i as \x using 0.65 + 2*\i] in {2,...,3} { % nodes at depth 2
    \node[tgpnode,fill=black!50!white] (d2\i) at (\x,-2) {$\times$};
}
\foreach \i [evaluate=\i as \cl using 2*\i, % drawing edges in between
             evaluate=\i as \cr using 2*\i + 1] in {0,...,3} {
    \draw (d2\i) -- (d3\cl.-270);
    \draw (d2\i) -- (d3\cr.-270);
}
\foreach \i [evaluate=\i as \x using 1.65 + 4*\i] in {0,...,0} { % nodes at depth 1
    \node[tgpnode] (d1\i) at (\x,-1) {$\times$};
}
\foreach \i [evaluate=\i as \x using 1.65 + 4*\i] in {1,...,1} { % nodes at depth 1
    \node[tgpnode,fill=black!50!white] (d1\i) at (\x,-1) {$\times$};
}
\foreach \i [evaluate=\i as \cl using 2*\i, % drawing edges in between
             evaluate=\i as \cr using 2*\i + 1] in {0,...,1} {
    \draw (d1\i) -- (d2\cl.-270);
    \draw (d1\i) -- (d2\cr.-270);
}
\node[tgproot] (r) at (3.65,0) {$\times$}; % the root
\draw (r) -- (d10.north);
\draw (r) -- (d11.north);
\end{scope}
\end{tikzpicture}
\caption{The optimal TGP program (right) of \maxproblemparam{$\{4\}$}{$\{\times,+\}$}{$\{t\}$} for $t\in[1,2)$. Gray internal nodes are 
%those can be 
discarded when converting to an equivalent CGP program (left) by 
%applying 
Lemma~\ref{lem:cgp-compactness}.} %using the identical subtrees.}
\label{fig:optimal-max-program}
\end{figure}
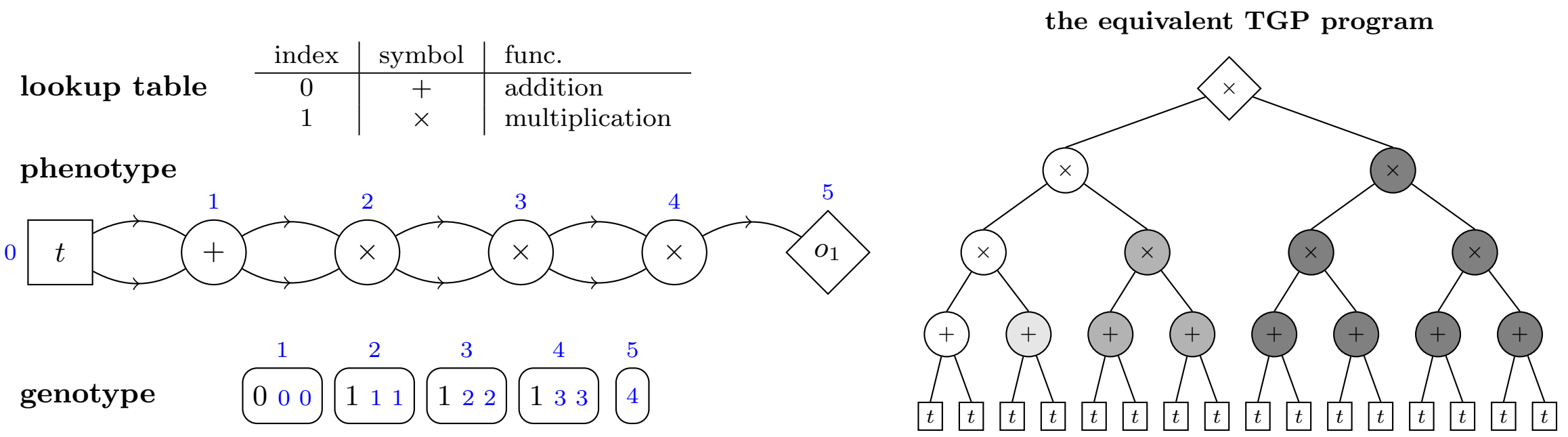

All bounds are exponential in $D$ since $n=2^{D+1}-1$, highlighting a key drawback of TGP, 
that its programs can be huge. 
Every TGP program has an equivalent CGP representation that can often be compressed. 
For \maxproblemfull, CGP uses 
    one input ($\Nin=1$ with $t>0$), 
    function set \newedit{$\{\times,+\}$} ($\Kfunc=2$, $\arity=2$), 
    $D$ function nodes \newedit{($\Nfunc=D$)}, and 
    one output \newedit{($\Nout=1$)}. 
\newedit{The setting of $\Nfunc$} 
%The bound on $\Nfunc=D$ 
follows by reducing an optimal symmetric TGP \newedit{program} via Lemma~\ref{lem:cgp-compactness}, 
\newedit{\eg see} Figure~\ref{fig:optimal-max-program} 
%shows the equivalent programs 
for $D=4$.
%\todo{Add the clarification that the genome length $L=\Nfunc(\arity+1)+\Nout$ is $3D+1$.}
\newedit{The genome length of CGP for \maxproblem is therefore $\genomelength=\Nfunc(\arity+1)+\Nout=3D+1$.}

We restrict our analysis to $t \geq 1$. In TGP,~\cite{KoetzingSNO2014} showed that 
$t$ does not affect $(1+1)$ TGP-single, while $(1+1)$ TGP-multi assumes $t=1$. For 
$t \in [1,2)$, in the optimal CGP program, the first node is addition and the rest 
multiplication, with sequential wiring to the output; for $t=2$ the first node may 
vary, and for $t>2$ it must be multiplication, otherwise the structure is unchanged.

\textbf{Notation in our analysis:}\label{sec:fitness-landscape}
%In our subsequent analyses, 
We often consider 
a specific CGP program in our analyses, \eg %usually 
the current solution of the ($1$+$1$)~CGP. When 
%the considered solution 
this  
is clear from the context, we then focus on its 
function nodes. 
%and use the following notation.
Function and input nodes are referred to directly by 
their node indices in the phenotype (\eg blue numbers in Figure~\ref{fig:optimal-max-program}). 
For a function node $i\in [D]$, we use $\fblkina{i}, \fblkinb{i}$ and $\fblkfunc{i}$ to 
denote the nodes (again by indices) that its first and second input edges are pointed 
to and its selected function of the lookup table. 
When we say setting $\fblkina{i}=j$ for $j<i$, we mean redirecting the first input of node $i$ to point 
at node $j$. The output value of node $i$ is denoted $\fblkval{i}$ 
and the domain of $\fblkval{i}$ is denoted $\fblkdom{i}$. The input $t$ serves as the base 
domain and value, \ie $\fblkdom{0}=\{t\},\fblkval{0}=t$.

%On the \maxproblemfull problem with $t\geq 1$, 
%In this section, we 
%We 
%first show that the maximum output of CGP programs for \maxproblemfull 
%with $t\geq 1$ is $\max\{2t,t^2\}^{2^{D-1}}$  
%by the following lemma. 
%This can be computed from the optimal TGP program, but can also be shown inductively from 
%the perspective of CGP as follows. 

\textbf{Maximum output:} %
The lemma below shows how the set $\fblkdom{i}$ grows as $i$ increases for \maxproblemfull 
with $t\geq 1$, and that the maximum output is $\max\{2t,t^2\}^{2^{D-1}}$ 
as $\max\{2t,t^2\}$ is the maximum of $\fblkdom{1}$, which is subsequently amplified
by the $D-1$ follow-up multiplications. %in an optimal program.

\begin{lemmarep}\label{lem:domain-fblk}
Consider any CGP program for \maxproblemfull with $t\geq 1$, $D\in\N$.
For any function node $i\in [D]$, the domain $\fblkdom{i}$ is defined by the following recurrence relation. 

\[
\fblkdom{i} = \begin{cases}
    \{t\} & \text{ if } i=0,\\
    %\left\{u+v, u\cdot v \;\bigg|\; u,v\in \bigcup_{k=0}^{i-1}\fblkdom{k}\right\} & \text{ if } 1\leq i \leq D.
    \{u+v, u\cdot v \mid u,v\in \bigcup_{k=0}^{i-1}\fblkdom{k}\} & \text{ if } 1\leq i \leq D.
    \end{cases}
\]
Consequently, it holds that: 
%\begin{enumerate}[label=\normalfont(\roman*)]
%%\item $\fblkdom{1}\subsetneq \fblkdom{2}\subsetneq \cdots \subsetneq \fblkdom{D} \subsetneq \fblkdom{D}$, %% only true if we include the 0
%\item $\max\{\fblkdom{i-1}\}< \max\{\fblkdom{i}\}$ for all $i\in[D]$, 
%\item $\min\{\fblkdom{i-1}\}\geq t$ for all $i\in[D+1]$, 
%\item the maximum output 
%%of a CGP program with $D\in \mathbb{N}$ function blocks 
%is
%%\[
%%\max\{\fblkdom{D}\} 
%%= \begin{cases}
%%    (2t)^{2^{D-1}} & \text{ if } 1\leq t\leq 2,\\
%%    (t^2)^{2^{D-1}}=t^{2^{D}} & \text{ if } t>2.
%%\end{cases}
%%\]
%$\max\{\fblkdom{D}\} = \max\{2t,t^2\}^{2^{D-1}}$.
%\end{enumerate}
{\upshape (i)} $\max\{\fblkdom{i-1}\}< \max\{\fblkdom{i}\}$ for all $i\in[D]$; 
{\upshape (ii)} $\min\{\fblkdom{i-1}\}\geq t$ for all $i\in[D+1]$; 
{\upshape (iii)} and the maximum output is $\max\{\fblkdom{D}\} = \max\{2t,t^2\}^{2^{D-1}}$.
\end{lemmarep}
\begin{appendixproof}
The recurrence relation holds because for each pair of  (not necessarily
distinct) values $u,v\in \bigcup_{k=0}^{i-1}\fblkdom{k}$, the function node $i$
offers at most two new values $u+v$ and $u\cdot v$. Statement (i) is also clear
because it suffices to set $\fblkfunc{i}=+$, $\fblkina{i}=i-1$ assuming
$\fblkval{i-1}$ has reached $\max\{\fblkdom{i-1}\}$ and set $\fblkinb{i}=0$.
This gives 
$\fblkval{i}=\fblkval{\fblkina{i}}+\fblkval{\fblkinb{i}}=\max\{\fblkdom{i-1}\}+t>\max\{\fblkdom{i-1}\}$, 
and implies $\max\{\fblkdom{i}\}>\max\{\fblkdom{i-1}\}$.

Statement (ii) can be shown by induction. The base case of $i=1$ trivially 
holds since $\fblkdom{0}=\{t\}$. Assume that this holds for all $k\in[i]$, 
that is, $\min\{\fblkdom{k-1}\}\geq t$, we need to show that $\min\{\fblkdom{i}\}\geq t$. 
By the recurrence relation, any $w\in\fblkdom{i}$ is either $u+v$ or $u\cdot v$ 
where $u\in \fblkdom{k_u}$ and $v\in \fblkdom{k_v}$ for some $k_u\in[i-1]$ 
and some $k_v\in[i-1]$. It follows from the induction hypothesise that $u\geq t$ 
and $v\geq t$, and this implies $u+v>t$ and $u\cdot v\geq t v \geq v \geq t$ 
given that $t\geq 1$. So, all $w\geq t$ and $\min\{\fblkdom{i}\}\geq t$.

We show (iii) by induction over $D$. The base case of $D=1$ holds because
$\fblkdom{1}=\{2t,t^2\}$, thus the maximum output is
$\max\{2t,t^2\}=\max\{2t,t^2\}^{2^{0}}$. For all $t\geq 1$, that output is at
least $2t\geq 2$, it then follows from (i) that for $D\in\N$,
$\max\{\fblkdom{i}\}\geq 2$ for all $i\in[D]$.
Assume that the maximum output holds for all $D\in [i]$, that is
$\max\{\fblkdom{i}\} = (\max\{2t,t^2\})^{2^{i-1}} =: w$, we need to show that
$\max\{\fblkdom{i+1}\} = (\max\{2t,t^2\})^{2^{i}} =
(\max\{2t,t^2\})^{2^{i-1}\cdot 2}=w^2$ with a new function node $i+1$. This
$w^2$-value can be produced by using the node $i+1$ with multiplication and make
sure that $\fblkina{(i+1)}=\fblkinb{(i+1)}=i$ assuming that $\fblkval{i}$ has
reached $w$ by the induction hypothesis. 
It remains to show that $w^2$ is the largest value of $\fblkdom{i+1}$. For all
pairs $u,v\in\bigcup_{k=1}^{i}\fblkdom{k}$, we get $w^2 =
\max\{\fblkdom{i}\}\cdot \max\{\fblkdom{i}\} \geq u\cdot v$ where the last
inequality follows from (i). Moreover, it follows from $w\geq 2$ as shown above
that $w^2 \geq 2 w = \max\{\fblkdom{i}\} + \max\{\fblkdom{i}\}\geq u + v$ and
here again we use (i). By the definition of $\fblkdom{i+1}$ in the recurrent
relation, these indeed imply $w^2=\max\{\fblkdom{i+1}\}$.
\end{appendixproof}

\begin{figure}[ht]\centering
\begin{tikzpicture}[scale=\ifarxiv1\else0.9\fi,
every node/.style={scale=0.9}]\small
\newcommand\nodeindex[1]{\scriptsize\textcolor{blue}{#1}}
\tikzset{middlearrow/.style={
                decoration={markings,
                mark= at position 0.5 with {\arrow{#1}} ,
            }, postaction={decorate} } }
\tikzstyle{block}=[rectangle, draw, rounded corners, minimum height=0.6cm]
\tikzstyle{node}=[circle, draw, fill=white, minimum size=0.7cm]
\tikzstyle{edge}=[]
\tikzstyle{tgpleaf}=[rectangle,draw,fill=white]
\tikzstyle{tgpnode}=[circle,draw,fill=white]
\tikzstyle{tgproot}=[diamond,draw,fill=white]

\begin{scope}[shift={(0.0,1.8)}] % phenotype
%\node[anchor=west] at (-1.0,0.8) {\textbf{phenotype}};
\node[node,rectangle,label=-180:{\nodeindex{$0$}}] (n0) at (0,0)    {$t$};
%\node[node,label=-180:{\nodeindex{$0$}}] (n0) at (0,-0.5) {$t$};
%\node[node,label=-180:{\nodeindex{$1$}}] (n1) at (0,0.5)  {$0$};
\node[node,label=90:{\nodeindex{$1$}}]   (n2) at (1.5,0)  {$+$};
\node[node,label=90:{\nodeindex{$2$}}]   (n3) at (3.0,0)  {$\times$};
\node[node,label=90:{\nodeindex{$3$}}]   (n4) at (4.5,0)  {$\times$};
\node[node,label=-90:{\nodeindex{$4$}}]   (n5) at (6.0,0)  {$+$};
\node[node,diamond,label=90:{\nodeindex{$5$}}]   (n6) at (7.5,0)  {$o_1$};

\draw[edge] (n0) edge[bend left,middlearrow={>}]  (n2);
\draw[edge] (n0) edge[bend right,middlearrow={>}] (n2);
\draw[edge] (n2) edge[bend left,middlearrow={>}]  (n3);
\draw[edge] (n2) edge[bend right,middlearrow={>}] (n3);
\draw[edge] (n3) edge[bend left,middlearrow={>}]  (n4);
\draw[edge] (n3) edge[bend right,middlearrow={>}] (n4);
\draw[edge] (n0) edge[bend left,middlearrow={>}]  (n5);
\draw[edge] (n0) edge[bend right,middlearrow={>}] (n5);
\draw[edge] (n4) edge[bend left,middlearrow={>}]  (n6);
\end{scope}
\end{tikzpicture}
\caption{A solution where at least two connection changes are required to increase the output value.}
\label{fig:non-monotone-landscape}
\end{figure}
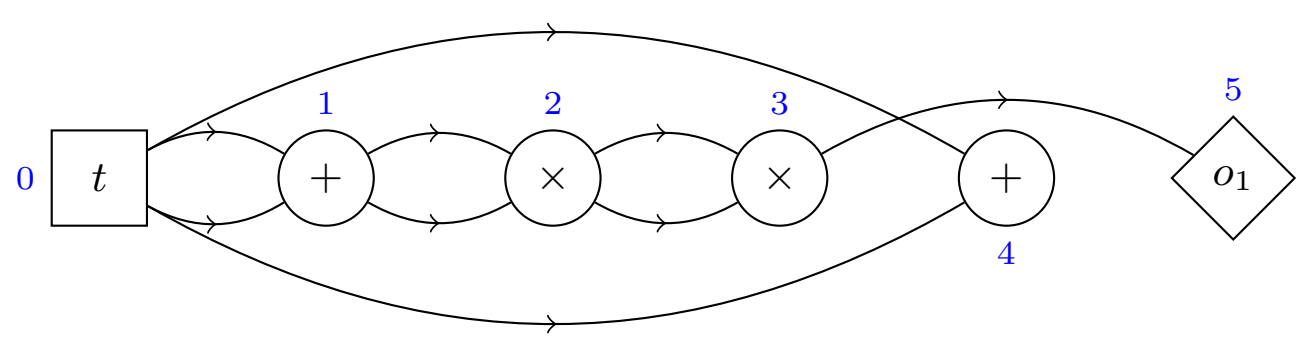

%Finally, we 
\textbf{Fitness landscape:} The fitness landscape of CGP on \maxproblemfull 
is not unimodal, that is, there are programs without a direct improved 
Hamming neighbor, \eg see 
%of Hamming distance~$1$. 
Figure~\ref{fig:non-monotone-landscape}. 
Thus the dynamics of CGP on \maxproblem is %can be 
%highly 
non-trivial.
\ifarxiv\else
Such programs can be used to derived lower bounds for CGP, but as being
non-essential to our message, those results are deferred to the appendix. 
\fi
%gives an example 
%of such a program. 

\begin{toappendix}
\ifarxiv
\section{Special Starts and Lower Bound}\label{sec:lower-bound}
\else
\section*{Omitted Section (Special Starts and Lower Bound)}
\addcontentsline{toc}{section}{\protect\numberline{}Omitted Section (Special Starts and Lower Bound)} 
\fi

%In this section, we 
We analyze the runtime of the ($1$+$1$)~CGP with SAM mutation 
and special initializations. 
%when it is initialized with some (nearly) ideal solutions. 
%, and subsequently give a basic 
%lower bound for the expected running time.
% 
We call a CGP program for \maxproblemfull \emph{structurally correct} if 
the output is connected to the last function node $D$, and for every function 
node $i\in[D]$, we have $\fblkina{i}=\fblkinb{i}=i-1$. That it, the program 
has the same graph structure as the optimal one. 
Moreover, provided that $t>1$, no change to the connections is 
possible without deteriorating the fitness. The structure of the graph is 
then correct and stable\footnote{For $t=1$, the program can still have the smallest output of $1$, thus the structure is not stable and can be easily destroyed.}. 
By a coupon-collecting argument, we can show the following 
expected time for SAM mutation to correct the function genes, one at a time 
as all function nodes are active. 

%Since only one change in each step can be made and we can also not revert any changes since this would decrease the fitness. We can also not change any connections, and also not the output node, since this also would decrease the fitness. Provided that $t > 1$}%, \ie modelling an 
%equivalent TGP program of a complete binary tree of depth $D$, however the 
%functions selected from the lookup table may not be optimal yet

\begin{theorem}\label{thm:sam-init-correct-struct}
The expected time for the ($1$+$1$)~CGP to optimize \maxproblemfull
for any $t>1$ using 
%the 
SAM mutation 
%operator 
and starting from 
a structurally correct program is 
%$(3+o(1))D\ln{D}=\Theta(D\log{D})$.
$\Theta(D\log{D})$.
\end{theorem}
\begin{proof}
%Since the initial solution has all function blocks structurally correct and the last function block $D+1$ is active, by Lemma~\ref{lem:sam-not-changing-x}, no connection genes of the current solution can be further modified by applying SAM mutation operator.  
For any $i\in[D]$, it follows from $\fblkina{i}=\fblkinb{i}=i-1$ that $\fblkval{i}$ 
is either $2\fblkval{i-1}$ or $\fblkval{i-1}^2$. 
By Lemma~\ref{lem:domain-fblk}~(ii), it holds that $\fblkval{i-1}\geq t>1$, thus
in both cases
%when $\fblkval{i}=2\fblkval{i-1}$ we get $\fblkval{i}>1 + \fblkval{i-1}>\fblkval{i-1}$
%and also when %$\fblkval{i}=\fblkval{i-1}^2$ 
$\fblkval{i}>\fblkval{i-1}$.
Thus $\fblkval{i}>\fblkval{i-1}$ for all $i\in[D]$, in other words, the connection 
genes are optimal and cannot be changed without deteriorating the output value 
unless the functions selected from the lookup table are changed in parallel. 
Note that all nodes are active so SAM only allows one change and parallel changes 
are disallowed.

Therefore, with probability $D/(3D+1) = 1/(3+o(1))$ a function gene is selected
and changed. Particularly, changing a function gene that is already optimized,
\ie matches the one in an optimal solution, also leads to the rejection of the
offspring. Therefore, we can model the progress using the following coupon
collecting process. First, we fix a specific optimal solution, then the coupons
are the function genes that are not yet correct with respect to that solution,
and in the worst case there can be $D$ of those. In each iteration, with
probability $1/(3+o(1))$ we run the standard coupon collector process, that is a
coupon arrive, and otherwise, no coupon arrives. It is known that the expected
time to collect all coupons in the standard process is $D H_D$ where $H_n$ 
denotes the $n$-th harmonic number. Thus the expected runtime of our process 
is $D H_D/(1/(3+o(1)))=\Theta(D\log{D})$.
\end{proof}

%Note that the above result does not hold for $t=1$ because a structurally correct
%solution in that case can still have the smallest output of $1$, hence the structure 
%can be easily destroyed. 
%\todo{Check what we can say about probabilistic mutation since parallel changes only occur with a probability of at most $1/(3D+1)^2$.}
%\andre{I think with probability $1-o(1)$ we can get the bound of $O(D \log(D))$. But the expected runtime bound could be much larger? I am not sure if we should expand on that too much. This $o(1)$ is just $O(1/\log(D))$.}
%\cuong{Yes, it only holds with probability and not in expectation. Also we need to be careful about how many two pairs that be changed there are.}
%\andre{Suggestion: Maybe switch this section with the section which presents the main results to "prepare" the reader?}

We now look at a program where it requires SAM mutation some time to make 
a change to its current solution. 
Since this is a suboptimal program, we have a basic lower bound for 
the expected runtime of the corresponding CGP. 
We call a CGP program \emph{missing the last} if the output 
is connected to the function node $D-1$, and for every function node $i\in[D-1]$, 
we have $\fblkina{i}=\fblkinb{i}=i-1$, and $\fblkfunc{i}=\times$ if $i\geq 2$ or $t\geq 2$
and otherwise $\fblkfunc{i}=+$, and $\fblkina{D}=\fblkinb{D}=0$. Figure~\ref{fig:non-monotone-landscape} 
gives such a program for $D=4$. 
To create a new program $y$ of non-decreasing fitness from such a suboptimal 
solution, it requires within a call to SAM mutation that at least one of $\fblkina{D},
\fblkinb{D}$ are redirected to node $D-1$, and then that the output is redirected to 
node $D$ in the last iteration of its while-loop. 
%Each of these events happens with 
%probability at most $\bigO(D^{-2})$, and 
Summing the probabilities of these events over all possible realizations of SAM mutation 
gives that the probability to produce $y$ is at most $\bigO(D^{-4})$. 
Thus, we obtain the following lower bound. 

\begin{theorem}\label{thm:sam-init-missing-last}
The expected time for the ($1$+$1$)~CGP 
%running on 
to optimize \maxproblemfull
for any $t\geq 1$ using SAM mutation and starting from a missing-the-last program 
%to modify (update) 
%its current program 
%that program 
is $\Omega(D^4)$.
\end{theorem}
\begin{proof}
For a lower bound result, it suffices to look at the waiting time to update the initial 
program (solution) to a different one. 
The initial solution is optimal for \maxproblemparam{$(D-1)$}{$\{\times,+\}$}{$\{t\}$}, 
thus to obtain a non-deteriorating solution by SAM mutation, the last function node $D$
must be turned active as follows. It requires that at least one of $\fblkina{D}$ 
and $\fblkinb{D}$ is changed to $D-1$ (instead of currently $0$), and that the unique 
change to an active node that triggers the exit of the while-loop in SAM connects the 
output to the last function node $D$ (instead of currently $D-1$). Such an outcome can only 
be produced if the while-loop runs for $i\geq 2$ iterations in which the first $i-1$ 
iterations only modify the inactive components $\{\fblkina{D},\fblkinb{D},\fblkfunc{D}\}$ 
and the last correctly modifies the output, thus with probability 
$\left(\frac{3}{3D+1}\right)^{i-1} \left(\frac{1}{(3D+1)D}\right)$. 

Conditioned the above event, we still need at least $\fblkina{D}$ or $\fblkinb{D}$ to be 
changed to $D-1$ during the first $i-1$ iterations, and here for an upper bound argument 
on the probability, we optimistically assume that having any correct changes among $i-1$ is 
sufficient (even though they may be overwritten by a wrong one later). Note that with probability $2/3$ 
we change $\fblkina{D}$ or $\fblkinb{D}$ (and not $\fblkfunc{D}$), then \textsc{NewValue} 
returns $D-1$ with probability of at most $1/(D-1)$ (after excluding the old value). 
%Conditioned on the event, the probability of having no correct changes during $i-1$ iterations
%is at most $\left(1-\frac{2}{3(D-1)}\right)^{i-1}$, thus in complement, the probability of 
%having at least one correct change is at most $1-\left(1-\frac{2}{3(D-1)}\right)^{i-1}\leq 
%\frac{2(i-1)}{3(D-1)}$ by Bernoulli's inequality. %% easier to argue this with a union bound
Thus conditioned on the event and by a union bound, the probability of having at 
least one correct change is at most $(i-1)\cdot \frac{2}{3(D-1)}$.
In summary, if the loop runs for 
$i\geq 2$ iterations, the probability of updating the current solution to a new one is at 
most $\left(\frac{3}{3D+1}\right)^{i-1} \left(\frac{1}{(3D+1)D}\right)\frac{2(i-1)}{3(D-1)}$. 
Summing this over all possible $i$ gives that the probability for SAM to update
the current solution is at most, here we use $\sum_{k=0}^{\infty}k a^k = a/(1-a)^2$:
\begin{align*}
&\sum_{i=2}^{\infty} \left(\frac{3}{3D+1}\right)^{i-1} \left(\frac{1}{(3D+1)D}\right)\frac{2(i-1)}{3(D-1)}\\
 \;\;&= \bigO(D^{-3}) \sum_{k=1}^{\infty} k\left(\frac{3}{3D+1}\right)^{k}
      = \bigO(D^{-3}) \cdot \frac{3/(3D+1)}{(1 - 3/(3D+1))^2}\\
 \;\;&= \bigO(D^{-3}) \cdot \frac{3}{3D+1} \cdot \frac{(3D+1)^2}{(3D-2)^2} = \bigO(D^{-4}).
\end{align*} 
Therefore, at least $\Omega(D^4)$ iterations are required in expectation to modify the current 
solution to a new one. This is also a lower bound on the optimization time as 
the initial solution is suboptimal.
\end{proof}
\end{toappendix}

\section{General Upper Bound}\label{sec:general-ub}

We now provide the main result of our paper, which is upper bounds for the expected runtime of 
($1$+$1$)~CGP on the \maxproblem problem for any initialization. 
The challenge in analyzing the dynamics of the current program while optimizing
the \maxproblem problem is to come up with a measure of the progress towards an
optimal program. It is particularly difficult to come up with a structural
measure because the output connection can be %modified by mutation. As this
% connection can
be moved around by mutation, \ie in favor of a fitness improvement elsewhere.
Hence, there is always a chance that \newedit{some nodes} of the evolved
structure so far become inactive and are destroyed in subsequent mutations.

To overcome this challenge, our strategy for showing an upper bound for the
expected runtime of CGP on \maxproblem is to consider fitness improvement. 
%We will explore in details in the next section that under all possible settings of the current program $x_0$,
In particular, if the algorithm is executed for a sufficiently long time, there
will be a good chance of a significant fitness improvement by a factor of
$\min\{a^{1/3},\max\{2t,t^2\}^{2^{D-1}}/a\}$ if the current fitness $f(x_0)$ 
is $a$.

\begin{lemma}
\label{lem:cgp-time-significant-imp}
%Consider the \CGPmax problem with the current output value $t<a<M$ with either pointwise mutation, or the \textsc{MutationSam} operator. Then the expected number of iterations the $(1+1)$-CGP algorithm needs to create a CGP program with output value $a^{4/3}$ or $M$ is $O(D^9)$.
Consider the ($1$+$1$)~CGP using SAM mutation to solve 
%the 
\maxproblemfull 
%problem 
for any $t\geq 1$ and $D\geq 2$. If the current 
program 
%(the best seen so far) 
$x_0$ has the output value $\max\{2t,t^2\}\leq f(x_0):=a <\max\{2t,t^2\}^{2^{D-1}}$, 
then there always exists a sequence of programs $x_0,\ldots,x_k$ with 
    $k=O(1)$, 
    $f(x_{i-1})\leq f(x_i)$ for all $i\in[k]$,
    $f(x_k)\geq \min\left\{\max\{2t,t^2\}^{2^{D-1}},a^{4/3}\right\}$,
and $\tilde{p}:=\prod_{i=1}^{k}\pmutsamlb(x_{i-1},x_i)$ with 
    $\tilde{p}=\Omega(D^{-8})$ if $t\in[1,2)$ and $\tilde{p}=\Omega(D^{-7})$ if $t\geq 2$.
    %that the SAM mutation operator can follow to increase the output of its current program significantly. 
%Consequently, 
Therefore, after at most $\bigO(D^8)$ iterations for $t\in[1,2)$ and $\bigO(D^7)$ 
for $t\geq 2$ in expectation, the output is either already maximized or 
at least $a^{4/3}$.
\end{lemma}

We will explore  
the proof of this key lemma %is given 
in the next section. Once proven, applying Lemma~\ref{lem:cgp-time-significant-imp} 
$\ell:=\lceil\frac{D-1}{\log(4/3)}\rceil = \bigO(D)$ times implies that 
the ($1$+$1$)~CGP using SAM mutation, and starting from a program with fitness 
$a\geq \max\{2t,t^2\}$, will reach a fitness of at least 
$a^{(4/3)^{\ell}}\geq \max\{2t,t^2\}^{2^{D-1}}$, \ie that of an optimal program,
in $\bigO(\ell\cdot D^8)=\bigO(D^9)$ and in $\bigO(\ell\cdot D^7)=\bigO(D^8)$ 
for $t\in [1,2)$ and $t\geq 2$ respectively. 
This is because 
    $(4/3)^{\ell} \geq (4/3)^{\frac{D-1}{\log(4/3)}} = \left((4/3)^{\log_{4/3}{2}}\right)^{D-1} = 2^{D-1}$ 
    and $a\geq 2$. 
%The expected number of fitness evaluations for this to happen is $\bigO(\ell\cdot D^8)=\bigO(D^9)$ 
%for $t\in [1,2)$ and $\bigO(\ell\cdot D^7)=\bigO(D^8)$ for $t\geq 2$. 
Reaching the initial fitness $\max\{2t,t^2\}$ is easy as it suffices to utilize 
the first function node, and the expected waiting time 
%for this is dominated by the aforementioned bounds. 
is dominated by the mentioned bounds. 
Those bounds are then transferred to that of the probabilistic
mutation by Lemma~\ref{lem:from-SAM-to-PROB}, and constitute our main results.

\begin{theoremrep}
\label{thm:cgp-upperbound}
%Consider the \CGPmax problem with either the pointwise mutation or the \textsc{MutationSam} procedure. Then the expected number of iterations the $(1+1)$-CGP algorithm needs to create a CGP program with output value $M$ is $O(D^{10})$.
The ($1$+$1$)~CGP using either \newedit{SAM mutation} or the probabilistic one
with parameter $p=1/(3D+1)$
and initialized with any program solves \maxproblemfull for any $t\in [1,2)$ in at most 
$\bigO(D^{9})$ iterations, and this bound is dropped to $\bigO(D^{8})$ for $t\geq 2$. 
\end{theoremrep}    
\begin{appendixproof}
%In case of pointwise mutation, to create an output value $a \geq 2t$, one can link both inputs of $N_0$ to the input node $0$ if necessary (with probability at least $p^2$), replace the multiplication at $N_0$ with an addition if $t = 1$ (with probability at least $p$), and finally connect the output node $N_o$ to $N_0$ if necessary (with probability at least $p / (D + 1)$). Together, these operations occur with probability at least $p^2 \cdot p/(D+1) \cdot p = \Omega(1/D^5)$. 
%\andre{In case of \textsc{MutationSam}, the operations in the order above are non decreasing in fitness. Hence, we obtain that with probability $\Omega(1/D^5)$ the desired fitness value is reached in four consecutive steps.}   
We first show the result for \newedit{SAM mutation}. The run of the algorithm
is divided into two phases: the first ends when the output of the current program 
is at least $\max\{2t,t^2\}$ and the second phase is completed when the optimal 
output is obtained. We first consider the case that $t\in[1,2)$.

To complete the first phase, it suffices to make sure that the first function node 
is optimized, \ie reaching its $\max\{\fblkdom{1}\}=\max\{2t,t^2\}$, and active. 
Once this value is reached
it is propagated, as the node is active, to the output either directly, 
or by multiplying or adding some other values of at least $t\geq 1$ by Lemma~\ref{lem:domain-fblk}~(ii) 
along the way. This guarantees that the output is at least $\max\{2t,t^2\}$, and can be 
achieved in at most 2 consecutive calls to SAM mutation as follows. 
The operator first corrects $\fblkfunc{1}$ if required (probability at least $1/(3D+1)$), 
then if required points the output to the first function node (probability at least 
$1/((3D+1)D)$). Note the first operation does not decrease the fitness and hence, it is feasible. 
The first phase therefore has a probability of $\Omega(1/D^3)$ to 
be completed per two iterations, thus it is done after $\bigO(D^3)$ iterations in expectation.

%Afterwards, the output of the program is $2t$, so the expected number of iterations until this occurs is $O(D^5)$. Then, by Lemma~\ref{lem:cgp-time-significant-imp}, for $\ell \in \mathbb{N}$, the expected number of iterations to obtain an output of at least $\min\{a^{(4/3)^\ell},M\}$ is $O(D^9 \cdot \ell)$. Note that for $\ell \geq (D - 1)/\log(4/3) = O(D)$, we have $(4/3)^\ell \geq 2^{D-1}$, and hence the output is at least $\min\{a^{(4/3)^\ell},M\} = M$, proving the theorem.
In the second phase, we apply Lemma~\ref{lem:cgp-time-significant-imp} to estimate the 
expected length of the phase. We assume that by the end of the first phase, the 
fitness of the current solution $x_0$ is at least $a:=f(x_0) \geq \max\{2t,t^2\}$, 
and let $M:=\max\{2t,t^2\}^{2^{D-1}}$ be the maximum output as shown in 
Lemma~\ref{lem:domain-fblk}~(iii). It follows from Lemma~\ref{lem:cgp-time-significant-imp} 
that after at most expected 
$\bigO(\ell\cdot D^{8})$ iterations, the fitness is increased to at least 
$\min\{a^{(4/3)^\ell},M\}$. By setting $\ell := \lceil \frac{D - 1}{\log(4/3)}\rceil = O(D)$, 
we get
\[
(4/3)^{\ell} \geq (4/3)^{\frac{D-1}{\log(4/3)}} 
= \left((4/3)^{\log_{4/3}{2}}\right)^{D-1} = 2^{D-1}.
\]
In other words, after at most $\bigO(\ell\cdot D^{8})=\bigO(D^{9})$ iterations, 
the optimal output $\min\{a^{(4/3)^\ell},M\}=M$ is reached.

Combining the two phases gives that the ($1$+$1$)~CGP using \newedit{SAM mutation} 
produces the optimal program after at most $\bigO(D^3)+\bigO(D^{9})$ iterations. 
The bound for $t\geq 2$ follows by the same proof except the expected length of 
the second phase, which dominates the runtime, is shortened to $\bigO(D^8)$.
% 
%\todo{Clarify that the genome length is $3D+1$ and recall the mutation rate parameter.}
\newedit{
The result for the probabilistic mutation follows from Lemma~\ref{lem:from-SAM-to-PROB} 
%and by noting that the estimates we have used satisfy the conditions of that lemma.
since we only use probability estimates from $\pmutsamlb$ and the mutation parameter is 
$p=1/(3D+1)=1/\genomelength$.}
\end{appendixproof}

\section{Time to a Significant Improvement}\label{sec:sig-improvement}

\newedit{We now present the proof of the key lemma (Lemma~\ref{lem:cgp-time-significant-imp}).
It requires a tedious case analysis and is split into several sublemmas corresponding to 
different settings of the initial solution. 
\ifarxiv\else
To convey the core idea, we will focus on 
explaining the case separation, and only present the complete proof for one critical 
setting (Lemma~\ref{lem:cgp-case-2-2-1}) while those for the remaining cases can be found 
in the appendix. 
\fi
We will state each setting directly and assume the same base conditions 
as in the key lemma. In particular, parameter $a$ is referred to as the fitness of the 
initial solution $f(x_0)$ and is assumed $a\geq \max\{t^2,2t\}$.} 

\newedit{
\ifarxiv
In the sublemmas, 
\else
The analysis of the critical setting demonstrates how the other proofs can be done. 
In those, 
\fi
further subcase divisions may be required, and in each
subcase, to argue about the probability of a significant improvement, \ie by a factor
of at least $a^{1/3}$, we often consider the worst case that some constant number
of calls to SAM mutation is required to achieve some step-by-step goal. For example,
a step may require to redirect an input connection of a specific function node to 
another specific node, \ie probability at least $1/(3D+1)$ to select the desired 
component in the encoded genotype then probability at least $1/(D-1)$
%\andre{Is this really this exact value? Do we need "at least" here? Depends on the connection we select! I also think that it is $1/(D+1)$? Look at the last node! Its connection has $D+1$ possibilities?} 
%\cuong{Good point! Added ``at least''. I think the bound $1/(D-1)$ is correct because even the last function node $D$ can only point its input connections to the preceding nodes $0,\ldots,D-1$. Additionally, Algorithm~\ref{algo:new-value} will force the exclusion of the old value (see lines 8-9 in that algorithm).}
to change its value correctly as Algorithm~\ref{algo:new-value} excludes the old value 
when sampling a new one, 
so probability $\Omega(D^{-2})$ for such a step. This can be improved to 
$\Omega(D^{-1})$ for redirecting to a function node located near the beginning, 
such as node $1$ or $2$, since 
%then this desired component can be selected with probability at least $\Omega(1)$}. % cuong: corrected this, the second probability is the one improved
making the correct change then occurs with a constant probability. 
In all cases, we only reply on lower bound estimates for $\pmutsamlb$.} 

\newedit{Proving Lemma~\ref{lem:cgp-time-significant-imp} requires 
identifying genome positions where large improvements occur. We begin with the easiest 
case where the last function node $D$ is inactive and can be used to increase the fitness.}

\begin{lemmarep}\label{lem:cgp-case-1}
If the function node $D$ is inactive then a single application of SAM mutation
has a probability of at least $\Omega(D^{-7})$ to increase the fitness by a factor
of $a$.
\end{lemmarep}
\begin{appendixproof}
Let $k<D$ be the node that the output is currently pointed to, \ie $\fblkval{k}=a$.
SAM only needs to make the following changes to the inactive components
(if required): setting $\fblkina{D}=\fblkinb{D}=k$ (probability 
$1/((3D+1)(D-1))^2$), and $\fblkfunc{D}=\times$ (probability $1/(3D+1)$).
Finally, it points the output to 
%function 
node $D$ (probability $1/((3D+1)D)$) and
this terminates 
%its main 
the while-loop. The fitness of the new solution is then 
$a^2$, and the probability of the event per call to SAM is $\Omega(D^{-7})$.
\end{appendixproof}

It is less trivial to identify a location for a significant improvement when node $D$ 
is active. We found that it is around the following node. Let $j^*$ be the first 
function node (\ie the smallest index) such that for all later function nodes $\ell>j^*$ 
(if they exist) it holds $\fblkina{\ell}=\fblkinb{\ell}=\ell-1 \wedge \fblkfunc{\ell}=\times$. 
The intuitive reason is that from this node, we can easily track the fitness. 
More generally, for any function node $\ell>j^*$ we have $\fblkval{\ell} = w^{2^{\ell-j^*}}$, 
where $w$ is the output value of node $j^*$. 
%The intuition is that for any function node $\ell>j^*$, $\fblkval{\ell}$ can
%be expressed as a function of $w:=\fblkval{j^*}$, that is, $\fblkval{\ell} = w^{2^{\ell-j^*}}$. $\ell>j^*$, $\fblkval{\ell}$ can
%be expressed as a function of $w:=\fblkval{j^*}$, that is, $\fblkval{\ell} = w^{2^{\ell-j^*}}$
%Thus 
%the final output of the current program 
%program output
%is $w^{2^{D-j^*}}$, particularly 
%we can observe the immediate impact to the final output from changing the input 
%values of $j^*$ and/or its function selection from the lookup table and further 
%break the cases. 
%\andre{Maybe also say that the output of the whole program for a given output of node $j^*$ is maximum and give the connection to the correct structure? To discuss!}
%\cuong{About your message about $j^*$ cannot be increased with SAM. I think it totally can, that is the thing with CGP: $j^*$ definition required that $D$ is active, thus it vanishes when $D$ is not. So imagine between SAM mutations where $j^*$ disappear and reappear again, it can totally be increased.}
%\andre{I agree :)}

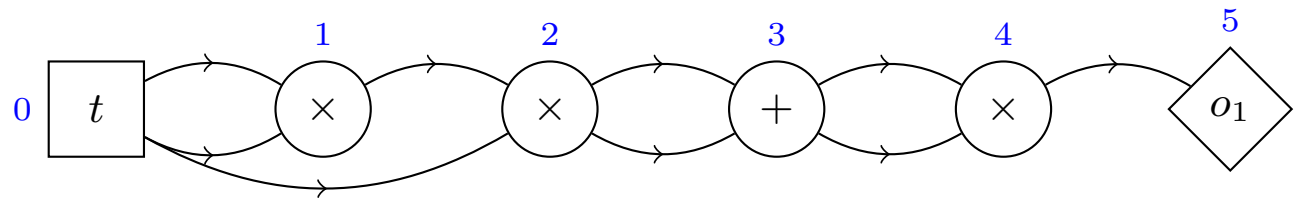
\begin{figure}[ht]\centering
\begin{tikzpicture}[scale=\ifarxiv1\else0.9\fi,
every node/.style={scale=0.9}]\small
\newcommand\nodeindex[1]{\scriptsize\textcolor{blue}{#1}}
\tikzset{middlearrow/.style={
                decoration={markings,
                mark= at position 0.5 with {\arrow{#1}} ,
            }, postaction={decorate} } }
\tikzstyle{block}=[rectangle, draw, rounded corners, minimum height=0.6cm]
\tikzstyle{node}=[circle, draw, fill=white, minimum size=0.7cm]
\tikzstyle{edge}=[]
\tikzstyle{tgpleaf}=[rectangle,draw,fill=white]
\tikzstyle{tgpnode}=[circle,draw,fill=white]
\tikzstyle{tgproot}=[diamond,draw,fill=white]

\begin{scope}[shift={(0.0,1.8)}] % phenotype
%\node[anchor=west] at (-1.0,0.8) {\textbf{phenotype}};
\node[node,rectangle,label=-180:{\nodeindex{$0$}}] (n0) at (0,0)    {$t$};
%\node[node,label=-180:{\nodeindex{$0$}}] (n0) at (0,-0.5) {$t$};
%\node[node,label=-180:{\nodeindex{$1$}}] (n1) at (0,0.5)  {$0$};
\node[node,label=90:{\nodeindex{$1$}}]   (n2) at (1.5,0)  {$\times$};
\node[node,label=90:{\nodeindex{$2$}}]   (n3) at (3.0,0)  {$\times$};
\node[node,label=90:{\nodeindex{$3$}}]   (n4) at (4.5,0)  {$+$};
\node[node,label=90:{\nodeindex{$4$}}]   (n5) at (6.0,0)  {$\times$};
\node[node,diamond,label=90:{\nodeindex{$5$}}]   (n6) at (7.5,0)  {$o_1$};

\draw[edge] (n0) edge[bend left,middlearrow={>}]  (n2);
\draw[edge] (n0) edge[bend right,middlearrow={>}] (n2);
\draw[edge] (n2) edge[bend left,middlearrow={>}]  (n3);
\draw[edge] (n0) edge[bend right,middlearrow={>}] (n3);
\draw[edge] (n3) edge[bend left,middlearrow={>}]  (n4);
\draw[edge] (n3) edge[bend right,middlearrow={>}] (n4);
\draw[edge] (n4) edge[bend left,middlearrow={>}]  (n5);
\draw[edge] (n4) edge[bend right,middlearrow={>}] (n5);
\draw[edge] (n5) edge[bend left,middlearrow={>}]  (n6);
\end{scope}
\end{tikzpicture}
\caption{An example solution where $j^*=3$.}
\label{fig:node-j-star}
\end{figure}

Figure~\ref{fig:node-j-star} gives an example on how to identify $j^*$. 
Note that by definition, if $\fblkfunc{D}=+$ then $j^*=D$. Otherwise, when 
$j^*$ is small enough, SAM can jump directly to the optimal program. 

\begin{lemmarep}\label{lem:cgp-case-2-1}
If function node $D$ is active and $j^*\leq 3$ then a sequence of $7$ applications 
of SAM mutation has a probability of at least $\Omega(D^{-7})$ to optimize \maxproblemfull.
\end{lemmarep}
\begin{appendixproof}
We consider the worst case of $j^*=3$ then the other cases follow as less work
is required for SAM (by skipping (ii)). The operator can make the following changes (if required) 
in order: 
    (i) setting $\fblkina{2}=\fblkinb{2}=1$ (probability $1/(3D+1)^2$); 
    (ii) setting $\fblkina{3}=\fblkinb{3}=2$ (probability $1/(2(3D+1))^2$);
    (iii) correcting $\fblkfunc{1}$, then $\fblkfunc{2}$ and finally $\fblkfunc{3}$ (probability $1/(3D+1)$ per correction).
    These steps (at most $7$) carrying out in that order do not create a program with 
    deteriorating fitness, hence possible immediate programs created (as we do not know 
    the active status of function nodes $1, 2$ and $3$), are always accepted by the algorithm. 
    We conclude that after at most $7$ calls to SAM, the probability to directly 
    optimize output is at least $\Omega(D^{-7})$.
\end{appendixproof}

For larger $j^*$, we need to break the cases based on how the node $j^*$
selects its function from the lookup table. Hereafter, we use the following notation for $j^*$: 
    $n_u:=\fblkina{j^*}$, 
    $n_v:=\fblkinb{j^*}$, 
    $u:=\fblkval{\fblkina{j^*}}$, and 
    $v:=\fblkval{\fblkinb{j^*}}$, 
and assume \wlo that 
$u\leq v$.

We first consider that an addition is selected, therefore, we have $\fblkval{j^*} = u + v$ 
and that $a 
    = \fblkval{D} 
    = (u+v)^{2^{D-j^*}} \leq (2v)^{2^{D-j^*}}
$. 
A subcase of this analysis, \ie $u,v$ are small, leads to a critical separation
in the result between $t\in[1,2)$ and $t\geq 2$.

\begin{lemma}\label{lem:cgp-case-2-2-1}
If function node $D$ is active, $j^*> 3$, and $\fblkfunc{j^*}=+$ then  
a sequence of $6$ applications of SAM mutation has a probability of at least 
$\Omega(D^{-8})$ to increase the fitness by a factor of $a^{1/3}$ if $t\in[1,2)$ and 
the probability is improved to $\Omega(D^{-5})$ if $t\geq 2$.
\end{lemma}
\begin{proof}
If $v\geq 4$, we argue that SAM only needs 
to perform the following changes (if required) in order: 
    (i) setting $\fblkina{j^*}=n_v$ (probability at least $1/((3D+1)(D-1))$);
    (ii) then changing $\fblkfunc{j^*}$ to $\times$ (probability $1/(3D+1)$).
These two steps in order cannot decrease the fitness, since the outputs of nodes $1,\ldots , (j^*-1)$ do not change while the outputs of nodes $j^*,\ldots,D$
cannot decrease. This requires at 
most 2 calls to 
%the operator 
SAM as node $j^*$ is active, and
the success probability is $\Omega(D^{-3})$. 

It follows from $v\geq 4$ that $v/2\geq (2v)^{1/3}$, so a success will lead to an 
improvement in fitness by a factor of at least
\begin{align*}
\frac{(v^2)^{2^{D-j^*}}}{a}
  &\geq \frac{\left(v^{2^{D-j^*}}\right)^2}{(2v)^{2^{D-j^*}}} 
   = \left(\frac{v}{2}\right)^{2^{D-j^*}}
   \geq ((2v)^{1/3})^{2^{D-j^*}} %\\
  %&= \left((2v)^{2^{D-j^*}}\right)^{1/3}
  %= b^{1/3}
   \geq a^{1/3}.
\end{align*}

Otherwise if $v<4$, and assume $t\in[1,2)$. Then it is sufficient for SAM to 
%make use 
use 
the two first function nodes $1$, $2$, 
%in conjunction with 
together with 
$j^*$ to increase the fitness. 
The changes (if required) in order are: %in order is:
    (i) setting $\fblkina{2}=\fblkinb{2}=1$ (probability $1/(3D+1)^2$); 
    (ii) correcting $\fblkfunc{1}$ to increase the output of $\fblkfunc{1}$ (probability $1/(3D+1)$) 
%    \andre{Why do we need to correct both here? Is it not enough to only correct the first one to a plus?}
%    \cuong{Fixed!}
%    \andre{good. But here comes the surprise: Suppose that the initial fitness is at least $2$. If both links of $N_0$ or the input $t$ are not connected to node $j^*$, then one other node $j'$ is connected to $j^*$. Then we can just carry over the analysis to that node $j'$ and we save a factor of $D$, since we already have the correct link to $j'$. If both are connected to the input or the first node, then the output of node $j^*-1$ is at most $2t^2$ (since we can assume $+$ in the first node). Then, just by changing $*$ to $+$ and the other way round, improves the output at node $j^*$ to $(2t)^2$.}
%    \cuong{It does not work, the ``carry over'' part, see the discussion document.}
    (iii) setting $\fblkina{j^*}=\fblkinb{j^*}=2$ (probability at least $1/((3D+1)(D-1))^2$)
          then setting $\fblkfunc{j^*}$ to $\times$ (probability $1/(3D+1)$).
%This requires 
Note that these steps in order do not decrease the fitness, since the outputs of all nodes do not decrease. Hence, those steps are feasible for SAM. At most $6$ calls to SAM are needed and the success probability is 
$\Omega(D^{-8})$.

If successful, we get $\fblkval{1}\geq 2$ thus $\fblkval{2}\geq 4$ regardless of 
whether $\fblkfunc{2}$ is $\times$ or $+$ and so $\fblkval{j^*}\geq 16$. 
%the new $\fblkval{j^*}\geq \max\{2t,t^2\}^{2^2}\geq (2t)^{2^2}\geq 16$ for $t\geq 1$. 
The fitness is then increased by a factor of at least
\begin{align*}
\frac{16^{2^{D-j^*}}}{(2v)^{2^{D-j^*}}}
  &> \frac{16^{2^{D-j^*}}}{8^{2^{D-j^*}}} 
   %= 2^{2^{D-j^*}}  
   = \left(8^{2^{D-j^*}}\right)^{1/3} 
   > \left((2v)^{2^{D-j^*}}\right)^{1/3}
   \geq a^{1/3}.%\qedhere
\end{align*}
Finally, when $t\geq 2$ we already have $\fblkval{1}\geq 2$ regardless of what 
$\fblkfunc{1}$ is. Thus it is sufficient for SAM to directly use function node $1$
to achieve the same fitness because $\fblkval{j^*}\geq 4^{2} \geq 16$. This 
increases the success probability to $\Omega(D^{-5})$.
\end{proof}

We now consider that $\fblkfunc{j^*}$ is $\times$, thus 
$a=(uv)^{2^{D-j^*}}\leq (v^2)^{2^{D-j^*}}$.  
Here if the preceding function node $j^{*}-1$ is inactive,
then the latter can be used to increase the fitness by a factor $a^{1/2}>a^{1/3}$.

\begin{lemmarep}\label{lem:cgp-case-2-2-2-1}
If function node $D$ is active but $j^{*}-1$ is not, and $j^*> 3$ 
and $\fblkfunc{j^*}=\times$, then a single application of SAM mutation has a probability 
of $\Omega(D^{-7})$ to increase the fitness by a factor of $a^{1/2}$.
\end{lemmarep}
\begin{appendixproof}
The following changes in order (if required) are sufficient 
for SAM:
    (i) setting $\fblkina{(j^*-1)}=\fblkinb{(j^*-1)}=n_v$ (probability at least $1/((3D+1)(D-2))^2$); 
    (ii) turning $\fblkfunc{(j^*-1)}$ to $\times$ (probability $1/(3D+1)$);
    (iii) finally setting $\fblkina{j^*} = j^{*}-1$ (probability $1/((3D+1)(D-1))$). 
Note that only the last change is for an active node, thus one call to SAM
is sufficient with a success probability of $\Omega(D^{-7})$.
If successful, the fitness is increased by a factor of
\begin{align*}
\frac{((v^2)\cdot v)^{2^{D-j^*}}}{a}
    &\geq \frac{(v^3)^{2^{D-j^*}}}{(v^2)^{2^{D-j^*}}}
    = v^{2^{D-j^*}} %\left(v^{3/2}\right)^{2^{D-j^*}}
    = \left((v^2)^{2^{D-j^*}}\right)^{1/2},
\end{align*}
which is at least $a^{1/2}$.
\end{appendixproof}

%Our final case considers $\fblkfunc{j^}=\times$ with $j^{}-1$ active 
Our final case considers \newedit{$\fblkfunc{j^*}=\times$ with function node $j^{*}-1$ active.}
%This completes 
%completing 
%the proof of the key lemma when combined with the previous cases in this section. 

\begin{lemmarep}\label{lem:cgp-case-2-2-2-2}
If nodes 
    $D, j^{*}-1$ are active, 
    $j^*> 3$, 
    and $\fblkfunc{j^*}=\times$, then a sequence of 4 applications of SAM mutation 
has a probability of $\Omega(D^{-7})$ to increase the fitness by a factor of $a^{1/3}$.
\end{lemmarep}
\begin{appendixproof}
It holds that exactly one input of $j^*$ must
be pointed to node $j^*-1$. This is firstly because by the definition of $j^*$, 
all nodes $\ell>j^*$ have used up all their inputs to the immediate preceding nodes, 
thus only the inputs of $j^*$ are available to make $j^*-1$ active. Secondly, it 
is not possible to have both inputs of $j^*$ pointing to $j^*-1$ because 
otherwise that contradicts
the definition $j^*$. 
%as the first node with the required property. 

In the case that $\fblkval{j^*-1}$ provides the input the smaller value 
for $j^*$, \ie 
$n_u = j^* - 1 \wedge n_v < j^* - 1$, it is sufficient for SAM to re-create
the same configuration of the solution as in the proof of the previous lemma. 
However, it requires less work but spans across multiple calls: 
    (i) setting $\fblkina{(j^*-1)}=\fblkinb{(j^*-1)}=n_v$ (probability at least $1/((3D+1)(D-2))^2$),
    and this will result in a non-deteriorating solution because 
    $\fblkval{\fblkina{(j^*-1)}},\fblkval{\fblkinb{(j^*-1)}}\leq u \leq v$; 
    (ii) finally turning $\fblkfunc{(j^*-1)}$ to $\times$ (probability $1/(3D+1)$) which does also not decrease the fitness. Thus the success probability is $\Omega(D^{-5})$ after at most $3$ calls to SAM, and if successful, by the same argument as in the previous lemma, the fitness is increased by a factor of $a^{3/4}$.

Otherwise if $\fblkval{j^*-1}$ is the input of the larger value of $j^*$, 
\ie $n_u < j^* - 1 \wedge n_v = j^* - 1$, we consider the two following subcases. 

If $v\geq u^2$: It is then sufficient for SAM to redirect $\fblkina{j^*}$ to  
%also 
$j^* - 1$, thus only one call to the operator has a success probability of 
$1/((3D+1)(D-1))=\Omega(D^{-2})$. This is because the fitness is then increased
by a factor of 
\begin{align*}
\frac{(v^2)^{2^{D-j^*}}}{(uv)^{2^{D-j^*}}}
  = \frac{(v^{2/3}\cdot v^{4/3})^{2^{D-j^*}}}{(uv)^{2^{D-j^*}}}
  \geq \frac{((u^2)^{2/3}\cdot v^{4/3})^{2^{D-j^*}}}{(uv)^{2^{D-j^*}}}
  = a^{1/3}.
\end{align*}

If $v<u^2$: The following changes can be done (if necessary):
    (i) redirecting $\fblkina{(j^*-1)}=\fblkinb{(j^*-1)}=n_u$ (probability at least $1/((3D+1)(D-2))^2$) which is feasible since we will only make this change if the incoming value is smaller than $u$;
    (ii) turning $(\fblkfunc{j^*-1})$ to $\times$ (probability $1/(3D+1)$) which is also feasible due to $u \geq 4$;
    (iii) redirecting $\fblkina{j^*}=j^*-1$ (probability $1/((3D+1)(D-1))$) which is also feasible, since $u^4 \geq u^3$.
So all these steps in order do not decrease the fitness and hence, are feasible for SAM. Thus at most 4 calls to SAM are required and the success probability is $\Omega(D^{-7})$ with an increase in fitness by a factor of at least
\begin{align*}
\frac{(u^4)^{2^{D-j^*}}}{(uv)^{2^{D-j^*}}}
  &= \frac{(u^{4/3}\cdot (u^2)^{4/3})^{2^{D-j^*}}}{(uv)^{2^{D-j^*}}}
   \geq \frac{(u^{4/3}\cdot v^{4/3})^{2^{D-j^*}}}{(uv)^{2^{D-j^*}}}
   = a^{1/3},
\end{align*}
and this concludes the proof.
\end{appendixproof}

\begin{figure}[h]\centering
\begin{tikzpicture}[scale=\ifarxiv1\else0.88\fi,
every node/.style={scale=0.9}]\small
\tikzstyle{block}=[rectangle, draw, rounded corners, minimum height=0.6cm]
\tikzstyle{edge}=[]

\node[block] (lem5)    at (0,0) {\textbf{Lemma~\ref{lem:cgp-time-significant-imp}} (all cases, assuming $t\geq 1$)};
\node[block] (lem7)    at (2,-1.25) {\textbf{Lemma~\ref{lem:cgp-case-1}}$\colon \Omega(D^{-7})$};
\node[block] (lem8-11) at (2,1.25) {Lemmas~\ref{lem:cgp-case-2-1}--\ref{lem:cgp-case-2-2-2-2}};

\draw[edge] (lem5) -- node[right] {$D$ inactive} (lem7);
\draw[edge] (lem5) -- node[left]  {$D$ active} (lem8-11);

\node[block] (lem8)    at (4.5,0) {\textbf{Lemma~\ref{lem:cgp-case-2-1}}$\colon \Omega(D^{-7})$};
\node[block] (lem9-11) at (4,2.5) {Lemmas~\ref{lem:cgp-case-2-2-1}--\ref{lem:cgp-case-2-2-2-2}};

\draw[edge] (lem8-11) -- node[right]  {$j^{*}\leq 3$} (lem8);
\draw[edge] (lem8-11) -- node[left] {$j^{*}> 3$} (lem9-11);

\node[block] (lem9)     at (7,1.25) {\textbf{Lemma~\ref{lem:cgp-case-2-2-1}}$\colon \Omega(D^{-8})$ if $t\in[1,2)$, $\Omega(D^{-5})$ if $t\geq 2$};
\node[block] (lem10-11) at (6,3.75) {Lemmas~\ref{lem:cgp-case-2-2-2-1}--\ref{lem:cgp-case-2-2-2-2}};

\draw[edge] (lem9-11) -- node[right] {$\fblkfunc{j^{*}}=+$} (lem9);
\draw[edge] (lem9-11) -- node[left]  {$\fblkfunc{j^{*}}=\times$} (lem10-11);

\node[block] (lem10) at (8.5,2.5) {\textbf{Lemma~\ref{lem:cgp-case-2-2-2-1}}$\colon \Omega(D^{-7})$};
\node[block] (lem11) at (8.5,5) {\textbf{Lemma~\ref{lem:cgp-case-2-2-2-2}}$\colon \Omega(D^{-7})$};

\draw[edge] (lem10-11) -- node[right] {$j^{*}-1$ inactive} (lem10);
\draw[edge] (lem10-11) -- node[left]  {$j^{*}-1$ active}   (lem11);
\end{tikzpicture}
\caption{\newedit{The case analysis constituting the proof of Lemma~\ref{lem:cgp-time-significant-imp} for $t\geq 1$. From left to right, the proof is branched into smaller sublemmas based on properties of the current program $x_0$ as described along the branching edges. Lower bounds on the probability of making a significant improvement by a factor of at least $(f(x_0))^{1/3}$ are written next to each sublemma.}
%\andre{Nice work! No mistake as far as I could figure out! I recommend to explain how this graph is to read! I guess from left to right somehow :). Roman, please also-recheck!}
%\cuong{Added some description.}
}
\label{fig:cgp-case-summary}
\end{figure}
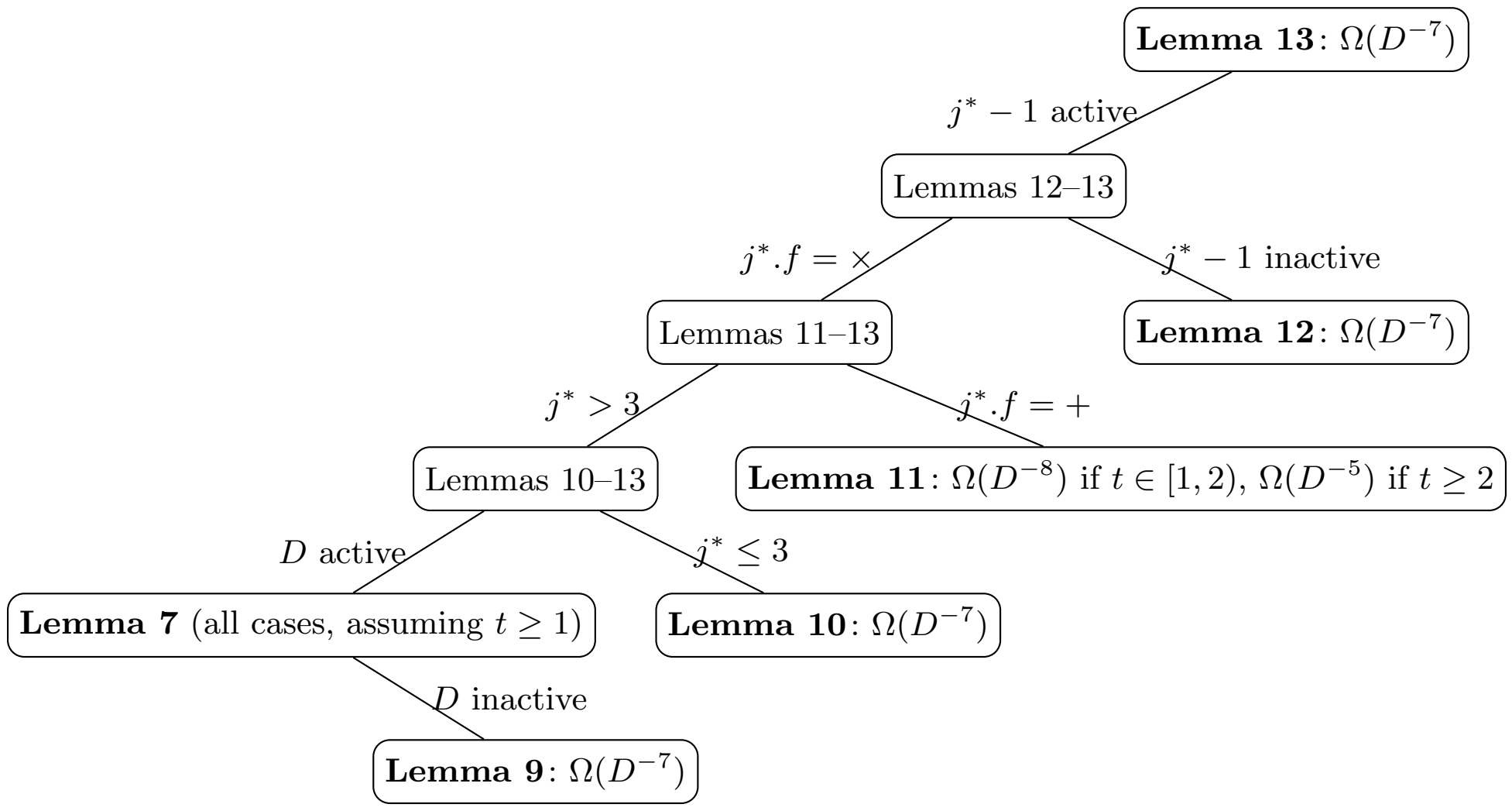

%The proof of the key lemma combines the ones in this section.
\newedit{The proof of the key lemma combines all the sublemmas as shown 
in the diagram of Figure~\ref{fig:cgp-case-summary}. The smallest 
improvement probability $\Omega(D^{-8})$ for $t<2$, \ie leading to bound 
$\bigO(D^9)$ in Theorem~\ref{thm:cgp-upperbound}, comes from a critical subcase
in Lemma~\ref{lem:cgp-case-2-2-1}. That subcase considers that node $D$ is active, 
node $j^*$ is located after node $3$ and is associated with $+$ in the
lookup table. However, if $j^*$ receives too low values from its two input
connections, \ie both less than $4$, then combining with $t<2$, a significant
fitness improvement requires in the worst case a precise use of nodes $1$ and
$2$ via node $j^*$, thus probability $\Omega(D^{-8})$. For $t\geq 2$, 
the setting of Lemma~\ref{lem:cgp-case-2-2-1} is no longer critical.}

\begin{proof}[Proof of Lemma~\ref{lem:cgp-time-significant-imp}]
The result follows from 
    Lemmas~\ref{lem:cgp-case-1}, 
           \ref{lem:cgp-case-2-1},
           \ref{lem:cgp-case-2-2-1},
           \ref{lem:cgp-case-2-2-2-1},
           and \ref{lem:cgp-case-2-2-2-2} \newedit{(\ie see Figure~\ref{fig:cgp-case-summary}).} 
\end{proof}

\section{Experiments}
\label{sec:experiments}

%\cuong{I removed all mentions of unsuccessful runs because, in my understanding, this is no longer the case after the bug with HVL-prime is fixed.} 

%In the framework of the empirical part of this work, we performed two types of experiments: The first experiment focused on a comparison between simple TGP and CGP to verify our theoretical findings and to study the efficiency of CGP on \maxproblem in comparison to TGP. With our second experiment we analyzed the performance of simple CGP for large settings of $D$ by using log-scaling. 
To accompany the theoretical findings, we performed two sets of experiments.
In the first, TGP and CGP were compared on 
%the \maxproblem problem 
\maxproblem with small $D$ 
to 
%illustrate 
show the exponential performance gain from CGP. In the second, we studied 
the empirical growth of the average runtime of CGP for larger $D$. 
%, and for this 
%a log-scaling technique is used to handle large fitness.

%\subsection{Experimental Setup}
\textbf{The setup:} 
Our code is written in Python and we used TinyverseGP~\cite{KalkreuthOJADVH2025}, a Python library with minimalistic implementations of various flavors of GP, to run the experiments. The $(1+1)$ TGP variants are denoted \texttt{simple\_tgp\_single} and \texttt{simple\_tgp\_multi}, using HVL-prime mutation as described in Section~\ref{sec:def-max} and~\cite{KoetzingSNO2014,KoetzingSNO2012}. The $(1+1)$ CGP variants are \texttt{simple\_\newedit{cgp}\_sam} and \texttt{simple\_\newedit{cgp}\_prob}, using SAM and \newedit{the} probabilistic mutation \newedit{respectively,} as in Algorithms~\ref{algo:mut-sam} and~\ref{algo:mut-pointwise}. The code and workflows for reproducing our results are publicly available\footnote{\url{https://github.com/GPTheory/NeurIPS26}}. 

%We evaluated various settings of $D\in\{1,\ldots,10\}$ and set $t=1$.
%For each value of $D$, 30 runs are performed with different random seeds per model (CGP and TGP) to ensure the statistical validity of our results. 
%The GP configuration used in our experiments is summarized in Table~\ref{tab:gp_configuration}.
%% 
%The runtime was measured as the number of iterations for the algorithm to evolve the optimal program, \ie the largest output $2^{2^{D-1}}$ is reached. 
%In our plots, we refer to 
%    the TGP models as \texttt{simple\_tgp\_single} and \texttt{simple\_tgp\_multi} and 
%    to the CGP models as \texttt{simple\_cgp\_sam} and \texttt{simple\_cgp\_prob}. 
%%We compared the number of iterations until the correct solution for \textsc{Max} was found. %% already said
%%To compare the performance of the models we calculated the average iterations per setting of $D$ and excluded the unsuccessful runs. 
%Statistics are reported per setting of $D$ excluding the unsuccessful runs. A run was declared unsuccessful when the maximum number of iterations was reached without a solution being obtained. 
For each experiment, $30$ runs were performed with different random seeds
and statistics are reported\newedit{. We also fixed an upper limit $2\times 10^6$ on 
the number of iterations allowed for each algorithm in each run, however
all runs terminated without reaching this.}
%attaining this limit.} 
%% -- note: we no longer have unsuccesful runs after fixing the bug spotted in Roman's implementation of HVL-prime
%excluding unsuccessful runs (only occurred with TGP).  
%A run was declared 
%unsuccessful if the maximum number of iterations, $2\times 10^6$, has passed
%without reaching the optimal program, \ie the one with the largest
%possible output. 
The GP configuration of our experiments is summarized in 
Table~\ref{tab:gp_configuration}. 
In the first set \newedit{of experiments}, we considered \maxproblemparam{$D$}{$\{\times,+\}$}{$\{1\}$}
for $D\in\{1,\ldots,10\}$ and compared TGP with CGP.
In the second set, we tested CGP on \maxproblemparam{$D$}{$\{\logmult,\logplus\}$}{$\{\log(2)\}$}
%so as to support
\newedit{to support} the large output of $D\in\{1,\ldots,30\}$ \newedit{as explained below.}
%and the 
%explanation is given below.

 \begin{table}[ht]\small
\centering 
%\scalebox{0.8}{
%\begin{tabular}{@{\extracolsep{4pt}}ll}
%\toprule   
% \textbf{Property} & \textbf{Setting} \\ 
%\midrule
%Depth $D$ & $\{1,\ldots,10\}$\\ 
%Input $t$ & $1$\\ 
%Function set & $\{+, \times\}$ \\ 
%Terminal set & $\{t\}$ \\ 
%Maximum iterations & $2000000$ \\ 
%Search strategy & ($1$+$1$) \\ 
%Selection type & non-strict \\ 
%No. replications per $D$ & 30 \\
%\bottomrule
%\end{tabular}
%}
%\scalebox{0.8}{
\begin{tabular}{@{\extracolsep{4pt}}llll}
\toprule   
 \textbf{\maxproblem param.} & \textbf{Setting} & 
 \textbf{GP param.} & \textbf{Setting} \\
\cmidrule{1-2}\cmidrule{3-4}
Depth $D$ & $\{1,\ldots,10\},\{1,\ldots,30\}$ & Maximum iterations & $2\times 10^6$\\%$2000000$ \\ 
Input $t$ & $1,2$ & \newedit{Search \& selection strategies} & \newedit{($1$+$1$), non-strict}\\ 
Function set & $\{\times,+\}$, $\{\logmult,\logplus\}$ & Number of replications per $D$ & 30\\ % & Selection type & non-strict\\ 
Terminal set & $\{t\}$, $\{\log{t}\}$ & \newedit{Mutation rate of prob. mutation} & $1 / (3D+1)$  \\ %& No. replications per $D$ & 30\\ 
%& & Mutation rate & $1 / (3D+1)$  \\ 
\bottomrule
\end{tabular}\vspace{0.3em}
%}
\caption{Configuration of GP and \textsc{Max} for the experiments.} 
\label{tab:gp_configuration}
\end{table}

\textbf{Log-scaling:} 
A challenge in conducting experiments on 
%the \textsc{Max} problem 
\maxproblemfull 
is
that the program output can be huge
%, \ie 
% as large as $\max\{2t,t^2\}^{2^{D-1}}$. 
%double exponential, 
as $D$ grows. 
%To conduct the experiments on moderate values of
%$D$, 
At some point, 
it is necessary to scale down the output and %. %by a $\log$-transform. 
% 
%To the
%evaluation process 
this can be done 
for $t>1$ 
%by using alternative functions for the lookup 
%table 
as follows. 
%here we assume $t>1$ 
%and describe the approach for 
Here we use $\log$ 
of base $2$ but other bases also work with minor adjustments.
%
%First, we define the $\logup\colon \R_0^+ \rightarrow \R_0^+$ function:
%\[
%\logup(x) := 
%    \begin{cases}
%    0 & \text{ if } x=0,\\
%    \log(x) & \text{ otherwise.}
%    \end{cases}
%\]
% 
%The down scaling is done by substituting 
We first substitute the input $t$ with $\log{t}$,  
then replace the $\mathrm{addition}(a,b)$ in the function table 
with\footnote{%If $|a-b|$ is still too large for the exponentiation 
          %%in the last case, 
          %we can use formula 
          A more general formula is $\max\{a,b\} 
          - C + \log(2^{C} + 2^{-|a-b|+C})$ 
          %instead, here 
          where $C>1$ can be 
          arbitrarily chosen.} 
$%\[
%\logplus(a,b) := 
%    \begin{cases}
%    a &\text{ if } b=0,\\
%    b &\text{ if } a=0,\\
%    a + b + \log(2^{-a}+2^{-b}) &\text{ otherwise},
%    \end{cases}
%% this is a simpler version
\logplus(a,b) := 
%    \begin{cases}
%    a &\text{ if } b=0,\\
%    b &\text{ if } a=0,\\
    \max(a,b) + \log(1 + 2^{-|a-b|}) %&\text{ otherwise},
%    \end{cases}
$ %\]
and the $\mathrm{multiplication}(a,b)$ with 
$%\[
\logmult(a,b) := 
%    \begin{cases}
%    0 &\text{ if } a=0 \vee b=0,\\
    a + b %&\text{ otherwise}.
%    \end{cases}
$. %\]
% 
%This will result in 
%a $\log$-transform being applied 
%to the output of every active node, and hence
%to the output of the program.
\newedit{The following lemma shows that these changes will result 
in a $\log$-transform being applied to the output of every program.}  
%and a more workable maximum $2^{D-1}\log(\max\{2t,t^2\})$.
The maximum output is then $2^{D-1}\log(\max\{2t,t^2\})$. 
%The consequence of these changes is that the $\logup$-transformation is applied 
%to the output of every active nodes of the solution. %\todo{Add a lemma about this}. %% -- no space  
%Hence, the optimal output is changed to 
%\begin{align*}
%\logup\left(\max\{2t,t^2\}^{2^{D-1}}\right)
%    &=\log\left(\max\{2t,t^2\}^{2^{D-1}}\right)\\
%    &=2^{D-1}\log\left(\max\{2t,t^2\}\right).
%\end{align*}
% 
Since the $\log$ function 
%is monotonically increasing, 
preserves the order of input values, %, \ie $a\geq
%b>0\Leftrightarrow \log{a}\geq \log{b}>0$, 
%the order between solutions with
%positive outputs remains unchanged. Moreover, if $t>1$ then the transformed
%output of any solution with a positive output is at least
%$\logup(t)=\log(t)>\log(1)=0=\logup(0)$, \ie better than that of any
%zero-output solution. 
%the order of programs with respect to their 
%output values is preserved under the transformation 
%$\logup$-transformation 
%for the whole search space. 
% if $t>1$. 
%So, 
the ($1$+$1$)~CGP behaves identically on 
%the transformed problem
\maxproblemparam{$D$}{$\{\logmult,\logplus\}$}{\newedit{$\{\log{t}\}$}} 
\newedit{as on the original problem} \maxproblemfull. %the original \maxproblem problem. 
%For this set of experiments, we use $t=2$ and allow $D$ up to $30$,
%the remaining parameters are the same as those in Table~\ref{tab:gp_configuration}.

\begin{lemmarep}\label{lem:log-transform}
\newedit{Let $f(x)$ be the output value of a CGP program encoded as a vector $x$ of integers 
and evaluated on \maxproblemfull, and $g(x)$ be the output value of the same encoded 
program but evaluated on \maxproblemparam{$D$}{$\{\log^{\times},\log^{+}\}$}{$\{\log{t}\}$}.  
%with matching indices in the lookup tables, \ie $\times$ for $\log^{\times}$ and $+$ for $\log^{+}$.
It holds that $g(x)=\log(f(x))$.} 
\end{lemmarep}
\begin{appendixproof}
\newedit{We will prove a more general result that the value, denoted $\fblkvalparam{i}{g}$, 
of any node indexed $i$ in the phenotype graph on 
\maxproblemparam{$D$}{$\{\log^{\times},\log^{+}\}$}{$\{\log{t}\}$}, 
regardless whether 
%this function node 
this node is active or not,
%of the input node
%(indexed $i=0$ in the phenotype graph) or of any function node (indexed $i\in [D]$),
%regardless whether this function node is active or not, 
is the $\log$-transform of the value, denoted $\fblkvalparam{i}{f}$, of the same node on 
the original problem \maxproblemfull. 
%\andre{Could we also provide this formulation in the statement of the lemma? Then say "In particular, we have that $g(x)=\log(f(x))$"?}
%\cuong{I tried to save space by avoid introducing the notation $\fblkvalparam{i}{f}$ in the main paper.}
That is, $\fblkvalparam{i}{g}=\log(\fblkvalparam{i}{f})$. The lemma then follows
because the program output is connected to one of these nodes
%, thus returning the value of that node.} 
and returns its value.} 
%\andre{Are we talking about function inputs or the values after calculation the function? I think we need the latter here! You can start the induction with $i=0$, because the calculation below also holds for every output of function node! But i am not happy with the formulation above! The proof is correct!}
%\cuong{OK, I saw the confusion. Added a clarification that those are the values of nodes in the phenotype graph. About the induction proof, we need to cover the case $i=0$ (and hence start with that as the base case) because a valid CGP program can still connect the program output directly to an input node. If we do not cover that case, then the sentence ``The lemma then follows because$\ldots$'' is incorrect.}

\newedit{The proof is done by induction on 
%the index $i$. 
$i$. The base case $i=0$ (the input node) 
holds by the definition of the terminal set in the transformed problem
\maxproblemparam{$D$}{$\{\log^{\times},\log^{+}\}$}{$\{\log{t}\}$}, that is,
$\fblkvalparam{0}{g}=\log{t}=\log(\fblkvalparam{0}{f})$. In the induction step,
we need to prove the result for (function) node indexed $i+1$ where $i\geq 0$, 
that is $\fblkvalparam{i+1}{g}=\log(\fblkvalparam{i+1}{f})$, assuming that 
this holds for all indices $j\leq i$. Let $a,b$ be the indices of the node(s) 
that the two input connections of node $i+1$ are currently connected to. By the 
design of CGP programs, we get $a\leq i$ and $b\leq i$, then it holds by 
the induction hypothesis (I.H.) that $\fblkvalparam{a}{g}=\log(\fblkvalparam{a}{f})$ 
and $\fblkvalparam{b}{g}=\log(\fblkvalparam{b}{f})$.}

\newedit{In the case that node $i+1$ is associated with $\times$ in the lookup 
table of the original problem, and hence $\log^{\times}$ in 
\maxproblemparam{$D$}{$\{\log^{\times},\log^{+}\}$}{$\{\log{t}\}$}, we have:
\begin{align*}
\fblkvalparam{i+1}{g} 
    &= \log^{\times}(\fblkvalparam{a}{g},\fblkvalparam{b}{g})
     \overset{\text{def}}{=} \fblkvalparam{a}{g} + \fblkvalparam{b}{g}
     \overset{\text{I.H.}}{=} \log(\fblkvalparam{a}{f}) + \log(\fblkvalparam{b}{f})\\
    &= \log(\fblkvalparam{a}{f} \times \fblkvalparam{b}{f}) 
     = \log(\fblkvalparam{i+1}{f}). 
\end{align*}
Otherwise if node $i+1$ is associated with $+$ in the lookup table of the 
original problem and so $\log^{+}$ in the transformed one, and assuming \wlo
that $\fblkvalparam{a}{g}\geq \fblkvalparam{b}{g}$, we get: 
\begin{align*}
\fblkvalparam{i+1}{g} 
    &= \log^{+}(\fblkvalparam{a}{g},\fblkvalparam{b}{g})
     \overset{\text{def}}{=} \max\{\fblkvalparam{a}{g},\fblkvalparam{b}{g}\}
       + \log(1+2^{-|\fblkvalparam{a}{g}-\fblkvalparam{b}{g}|})\\
    &= \fblkvalparam{a}{g} + \log(1+2^{-(\fblkvalparam{a}{g}-\fblkvalparam{b}{g})})
     = \log\left(2^{\fblkvalparam{a}{g}} \left(1+2^{-(\fblkvalparam{a}{g}-\fblkvalparam{b}{g})}\right)\right)\\
    &= \log\left(2^{\fblkvalparam{a}{g}}+2^{\fblkvalparam{b}{g}}\right)
     \overset{\text{I.H.}}{=} \log\left(2^{\log(\fblkvalparam{a}{f})}+2^{\log(\fblkvalparam{b}{f})}\right)\\
    &= \log(\fblkvalparam{a}{f}+\fblkvalparam{b}{f}) = \log(\fblkvalparam{i+1}{f}).
\end{align*}
Combining the two cases completes the induction step and hence the proof.}
\end{appendixproof}
\begin{toappendix}
\newedit{Although $\log$-transform was not needed in our experiments with TGP, 
we note that the statement of the lemma also holds for TGP programs. The proof 
of this follows the same steps as in the above proof, but the induction is done 
on the height of TGP subprograms.}\par  
\end{toappendix}

\begin{figure}
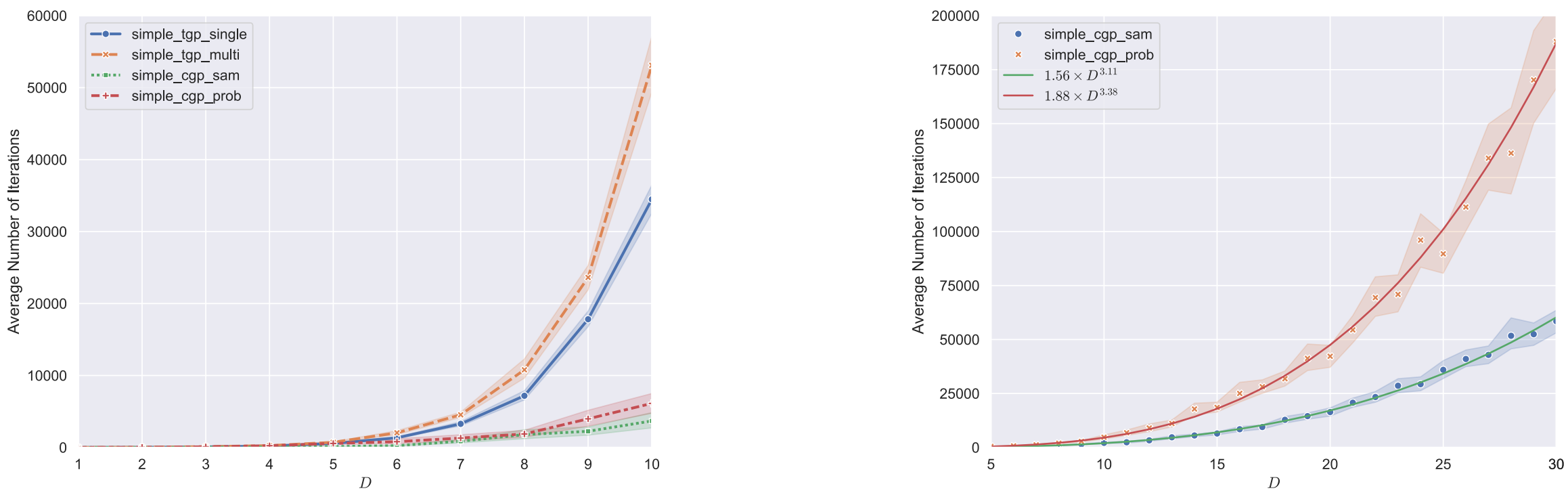
\centering
\begin{subfigure}[b]{0.48\textwidth}\centering
    \includesvg[scale=0.33]{figures/cr_revised/max_plus_mul_tgp_cgp_1_10.svg}
    \caption{TGP and CGP on \maxproblemparam{$D$}{$\{+,\times\}$}{$\{1\}$}.}
    \label{fig:experiments-cgp-tgp}
\end{subfigure}\hfill
\begin{subfigure}[b]{0.48\textwidth}\centering
    \includesvg[scale=0.33]{figures/cr_revised/max_plus_mul_log_cgp_sam_prob_5_30_logloglinfit.svg}
    \caption{CGP on \maxproblemparam{$D$}{$\{\logplus,\logmult\}$}{$\{\log(2)\}$}.}
    \label{fig:experiments-cgp-log-scale}
\end{subfigure}
\caption{Empirical results on \maxproblem problems comparing TGP and CGP for small $D$ (left), and testing CGP for larger $D$ (right). 
Results for too small $D$ are dropped in the right figure to reduce noise when applying linear regression on the log-log scale to deduce the growth orders.} 
%(red and green curves).}
\label{fig:experiments}
\end{figure}

%\subsection{Results}
\textbf{The results:} 
%%In addition to the plots presented in Figure~\ref{fig:cgp_tgp_comparison}, 
%Figure~\ref{fig:cgp_tgp_comparison} shows the results for our comparison between the tested TGP and CGP models which confirm the theoretical results presented in previous sections and shown in~\cite{KoetzingSNO2014, KoetzingSNO2012}. While the average runtime for the TGP models grows exponentially when the setting of $D$ is gradually increased, it is clearly visible that there is a stark contrast in terms of runtime for the considered TGP and CGP models, which aligns with our theoretical results above. Please note that 
%we did not obtain any successful runs for \texttt{simple\_tgp\_single} on $D=10$. 
%%The corresponding 
%The statistics are provided in the appendix document.
Figure~\ref{fig:experiments-cgp-tgp} shows our results for the first set of experiments,
comparing TGP and CGP. It is clearly visible that there is a stark contrast in
terms of runtime between TGP and CGP, particularly, the runtime of TGP grows
extremely large as $D$ increases. This aligns with our theoretical results and 
with those of~\cite{KoetzingSNO2014, KoetzingSNO2012}. 
%Note that we did
%not obtain any successful runs for \texttt{simple\_tgp\_single} on $D=10$. 
Additional statistics are provided in the \newedit{repository}.

Figure~\ref{fig:experiments-cgp-log-scale} shows the results for the second set of log-scaling experiments.  
%which demonstrates that 
We observe that CGP is able to obtain solutions for \textsc{Max} even with large settings of $D$ in a relatively small period of time when compared to the search performance of TGP %on \textsc{Max} 
in Figure~\ref{fig:experiments-cgp-tgp}. 
%Our second experiments also demonstrates the some performance gap between probabilistic mutation and SAM on \maxproblem.
The results suggest that SAM mutation is faster than the probabilistic mutation on \maxproblem.
We also deduced the growth orders by applying linear regression to the average runtime in the log-log scale (applying $\log$ to both axes). The observed orders (red and green curves in Figure~\ref{fig:experiments-cgp-log-scale}) are between $\Omega(D^3)$ and $\bigO(D^4)$. They agree with our theoretical bound $\bigO(D^8)$ for $t=2$ but suggest that CGP may run faster in practice.

%\section{Discussion}\label{sec:discussion}
%\romann{We could address the work from a perspective of GP practitioner here by reflecting/discussing on what the results mean for the empirical and application level to create some connecting points to previous empericical work.}

\section{Conclusions}\label{sec:conclusion}

We presented the first rigorous runtime analysis of CGP on the \maxproblemfull problem, a well established TGP benchmark whose runtime for mutation-only TGP is known to be superlinear in the optimal program length $n = 2^{D+1}-1$. We showed that the $(1+1)$ CGP algorithm, using either \newedit{the} probabilistic mutation or SAM, finds an optimal program in expected time $\bigO(\log^9 n)$, yielding an exponential speedup over TGP on this problem. The analysis also identifies how SAM \newedit{mutation} can exploit specific genome locations to obtain large fitness gains. Our theoretical findings are supported by experiments on small depths ($D \leq 10$), which already show a substantial gap between TGP and CGP, and extendable to larger $D$ via log-scaling. 
Our upper bounds may not be tight as experiments suggest CGP runs faster in practices, but they are sufficient for showing the exponential gap. Our proof requires tedious case analysis, but this was expected for the first step exploring the new search space. We are optimistic that it can be improved in the future. 
%
%As runtime analysis has significantly advanced our understanding of TGP~\cite{KoetzingSNO2014,LissovoiO2018}, it is natural to extend similar techniques to other GP variants such as CGP. 
We hope this work provides a useful step toward\newedit{s a better understanding 
%different flavors of GP. %that direction.
of GP representations for programs.}
%\andre{cheese and wine gets nearer :D. I really like your changes, Cuong :). Next time I really watch out with ChatGPT.}

\ifarxiv
\bigskip\noindent\textbf{Acknowledgements.} 
\newedit{The authors would like to thank Fabricio Olivetti de Fran\c{c}a who pointed out to us that in an earlier version of this work, we analyzed a more complex form of the \maxproblem problem than the one studied in the literature, and Jakob Bossek for the discussion prior to this work about GP literature. Roman Kalkreuth is supported by an Alexander von Humboldt Professorship in AI held by Holger Hoos.}
\else
\begin{ack}
\newedit{The authors would like to thank Fabricio Olivetti de Fran\c{c}a who pointed out to us that in an earlier version of this work, we analyzed a more complex form of the \maxproblem problem than the one studied in the literature, and Jakob Bossek for the discussion prior to this work about GP literature. Roman Kalkreuth is supported by an Alexander von Humboldt Professorship in AI held by Holger Hoos.}
\end{ack}
\fi

\bibliographystyle{abbrvnat}
\bibliography{neurips26,neurips26_more} % _more file is for not-yet-standardized items

%% -- TECHNICAL APPENDIX AND SUPPLEMENTARY MATERIALS
%\clearpage %% can be renable if there is something else to discuss technically
\appendix

%% -- SUBMISSION CHECK-LIST (see the end of Conclusions)
\ifarxiv
\else
\AtEndDocument{
\newpage
\section*{NeurIPS Paper Checklist}
\addcontentsline{toc}{section}{\protect\numberline{}NeurIPS Paper Checklist}

\begin{enumerate}

\item {\bf Claims}
    \item[] Question: Do the main claims made in the abstract and introduction accurately reflect the paper's contributions and scope?
    \item[] Answer: \answerYes{}. %\answerTODO{} % Replace by \answerYes{}, \answerNo{}, or \answerNA{}.
    \item[] Justification: Our abstract and introduction claim a proof for the performance guarantees of Cartesian Genetic Programming (CGP) on the \maxproblem problems, showing an exponential speed-up compared to Tree-based GP (TGP). The result sheds the light on the benefit of the graph-based representation that CGP brings and our analysis shows how CGP can navigate its search and fitness spaces efficiently.
    %\justificationTODO{}

\item {\bf Limitations}
    \item[] Question: Does the paper discuss the limitations of the work performed by the authors?
    \item[] Answer: \answerYes{}. %\answerTODO{} % Replace by \answerYes{}, \answerNo{}, or \answerNA{}.
    \item[] Justification: We discussed the limitations of our results in the conclusions. While our upper bounds may not be tight, as experiments suggest CGP may run faster in practice, they are sufficient for demonstrating the exponential speedup. Note that experiments can only explore limited values for $D$, while the theoretical result holds asymptotically as $D$ grows to infinity. Our proof requires tedious case separation, a necessary step for 
    establishing a first result in a new territory. Despite these challenges, we remain optimistic that these can be improved upon in future work.    
    %\justificationTODO{}

\item {\bf Theory assumptions and proofs}
    \item[] Question: For each theoretical result, does the paper provide the full set of assumptions and a complete (and correct) proof?
    \item[] Answer: \answerYes{}. 
    %\answerTODO{} % Replace by \answerYes{}, \answerNo{}, or \answerNA{}.
    \item[] Justification: Our main result (Theorem~\ref{thm:cgp-upperbound}) is clearly stated with all necessary assumptions. The algorithms, operators and benchmark function it covers are precisely defined in Sections~\ref{sec:def-cgp} and~\ref{sec:def-max}. The necessary lemmas constructing its proof are stated in the paper, and their proofs along the proof of the main result are fully given in the appendix document. Due to space restriction, some results non-essential for our message are deferred to the appendix and their proofs are given there. 
    %\justificationTODO{}

    \item {\bf Experimental result reproducibility}
    \item[] Question: Does the paper fully disclose all the information needed to reproduce the main experimental results of the paper to the extent that it affects the main claims and/or conclusions of the paper (regardless of whether the code and data are provided or not)?
    \item[] Answer: \answerYes{} % Replace by \answerYes{}, \answerNo{}, or \answerNA{}.
    \item[] Justification: We have described the experimental setup in detail, disclosed the source code that was used for the experiments and provided the plot scripts to reproduce the figures. 

\item {\bf Open access to data and code}
    \item[] Question: Does the paper provide open access to the data and code, with sufficient instructions to faithfully reproduce the main experimental results, as described in supplemental material?
    \item[] Answer: \answerYes{} % Replace by \answerYes{}, \answerNo{}, or \answerNA{}.
    \item[] Justification: The code used for the experiments builds upon a open source benchmarking tool. The link to the modified, anonymized version that has been used for this specific use case, is provided in the paper. The execution scripts are also part of the content of the repository. The benchmarking tool used is properly documented and test scripts are provided to exemplify its usage. 

\item {\bf Experimental setting/details}
    \item[] Question: Does the paper specify all the training and test details (e.g., data splits, hyperparameters, how they were chosen, type of optimizer) necessary to understand the results?
    \item[] Answer: \answerYes{} % Replace by \answerYes{}, \answerNo{}, or \answerNA{}.
    \item[] Justification: The problem considered does not require a training and test data split. However, we have listed the settings of the hyperparameters. Please note, since we only considered simple models for our experiments used to complement theoretical findings, we did not perform hyperparameter optimization.  

\item {\bf Experiment statistical significance}
    \item[] Question: Does the paper report error bars suitably and correctly defined or other appropriate information about the statistical significance of the experiments?
    \item[] Answer: \answerYes{} % Replace by \answerYes{}, \answerNo{}, or \answerNA{}.
    \item[] Justification: Confidence intervals (CIs) are included in our plots, and additional statistics are provided in the appendix document. Our empirical results indicate a clear performance gap between CGP and TGP, and there are no overlaps of CIs for $D$ that we are interested in, so significance tests are not required. We emphasize that our main contribution is theoretical for the runtime of CGP in expectation, and the experiments were conducted to only complement the theoretical findings.  

\item {\bf Experiments compute resources}

%CPU Partition
%Consisting of 17 nodes with 2 AMD EPYC 7543 2.8GHz. The first
%four nodes have 32x64GB of RAM, the rest 16x64GB

% 30 single-core jobs ran in parallel with 3GB of maximum memory capacity per job to run one instance per setting of D 

% Max runtime for each job 24 hours -> max time for each experiment 

    \item[] Question: For each experiment, does the paper provide sufficient information on the computer resources (type of compute workers, memory, time of execution) needed to reproduce the experiments?
    \item[] Answer: \answerYes{} % Replace by \answerYes{}, \answerNo{}, or \answerNA{}.
    \item[] Justification: The general experimental setup is described in detail in the experiments section. The source code is publicly available and the repository also contains the execution scripts. In addition we report technical setup that was used to perform the experiments: 
    \begin{itemize}
        \item CPU Partition of our local (non-public) compute cluster 
        \begin{itemize}
            \item 18 CPU Nodes with 2x AMD EPYC 7543 32-Core Processor, the first four nodes have 32x64GB of RAM, the rest 16x64GB
            \item 30 single-core jobs with 3GB of maximum memory capacity ran in parallel per setting of $D$
            \item Maximum runtime for each job was limited to 24 hours. The maximum runtime for each experiment was therefore also 24 hours. 
            \item We performed a round of preliminary experiments, with a lower sample size to test the implementations and general setup. 
        \end{itemize}
    \end{itemize}
    %\justificationTODO{}
    
\item {\bf Code of ethics}
    \item[] Question: Does the research conducted in the paper conform, in every respect, with the NeurIPS Code of Ethics \url{https://neurips.cc/public/EthicsGuidelines}?
    \item[] Answer: \answerYes{}.%\answerTODO{} % Replace by \answerYes{}, \answerNo{}, or \answerNA{}.
    \item[] Justification: We have read and strictly follow the NeurIPS Code of Ethics.%\justificationTODO{}

\item {\bf Broader impacts}
    \item[] Question: Does the paper discuss both potential positive societal impacts and negative societal impacts of the work performed?
    \item[] Answer: \answerNA{}. %\answerTODO{} % Replace by \answerYes{}, \answerNo{}, or \answerNA{}.
    \item[] Justification: Our work is purely theoretical, there are no foreseeable negative societal impacts as we know of. The main contribution of our paper is to advance our understanding of search spaces of programs, beyond the conventional tree-based representation, in GP. %\justificationTODO{}
    
\item {\bf Safeguards}
    \item[] Question: Does the paper describe safeguards that have been put in place for responsible release of data or models that have a high risk for misuse (e.g., pre-trained language models, image generators, or scraped datasets)?
    \item[] Answer: \answerNA{}. %\answerTODO{} % Replace by \answerYes{}, \answerNo{}, or \answerNA{}.
    \item[] Justification: The paper poses no such risks. %\justificationTODO{}

\item {\bf Licenses for existing assets}
    \item[] Question: Are the creators or original owners of assets (e.g., code, data, models), used in the paper, properly credited and are the license and terms of use explicitly mentioned and properly respected?
    \item[] Answer: \answerYes{}. %\answerTODO{} % Replace by \answerYes{}, \answerNo{}, or \answerNA{}.
    \item[] Justification: We used the TinyverseGP library for our experiments, cited the original paper, and provided the required license in our public repository for our code and experiments. 
    %\justificationTODO{}

\item {\bf New assets}
    \item[] Question: Are new assets introduced in the paper well documented and is the documentation provided alongside the assets?
    \item[] Answer: \answerNA{}. %\answerTODO{} % Replace by \answerYes{}, \answerNo{}, or \answerNA{}.
    \item[] Justification: The paper does not release new assets. We rigorously analyzed an existing search paradigm in GP. %\justificationTODO{}

\item {\bf Crowdsourcing and research with human subjects}
    \item[] Question: For crowdsourcing experiments and research with human subjects, does the paper include the full text of instructions given to participants and screenshots, if applicable, as well as details about compensation (if any)? 
    \item[] Answer: \answerNA{}. %\answerTODO{} % Replace by \answerYes{}, \answerNo{}, or \answerNA{}.
    \item[] Justification: The paper does not involve crowdsourcing nor research with human subjects.
    %This is a theory paper. Chalks and sharpies cannot read on their own anyway. %\justificationTODO{}

\item {\bf Institutional review board (IRB) approvals or equivalent for research with human subjects}
    \item[] Question: Does the paper describe potential risks incurred by study participants, whether such risks were disclosed to the subjects, and whether Institutional Review Board (IRB) approvals (or an equivalent approval/review based on the requirements of your country or institution) were obtained?
    \item[] Answer: \answerNA{}. % \answerTODO{} % Replace by \answerYes{}, \answerNo{}, or \answerNA{}.
    \item[] Justification: The paper does not involve crowdsourcing nor research with human subjects.
    %This is a theory paper. Only chalks and sharpies were harmed in the process. %\justificationTODO{}

\item {\bf Declaration of LLM usage}
    \item[] Question: Does the paper describe the usage of LLMs if it is an important, original, or non-standard component of the core methods in this research? Note that if the LLM is used only for writing, editing, or formatting purposes and does \emph{not} impact the core methodology, scientific rigor, or originality of the research, declaration is not required.
    %this research? 
    \item[] Answer: \answerNA{}. %\answerTODO{} % Replace by \answerYes{}, \answerNo{}, or \answerNA{}.
    \item[] Justification: The core method development in this research does not involve LLMs as any important, original, or non-standard components.
    %We only use LLMs to polish our writing. We are not fond of them altering our writing style, but they are fun to interact with and sometimes write non-sense.%\justificationTODO{}

\end{enumerate}
}
\fi

%\iffinal\else
%\section{Results}
%\label{sec:app_results}
%\fi
\end{document}